\documentclass[letterpaper]{article} % DO NOT CHANGE THIS
\usepackage{aaai25}  % DO NOT CHANGE THIS
\usepackage{times}  % DO NOT CHANGE THIS
\usepackage{helvet}  % DO NOT CHANGE THIS
\usepackage{courier}  % DO NOT CHANGE THIS
\usepackage[hyphens]{url}  % DO NOT CHANGE THIS
\usepackage{graphicx} % DO NOT CHANGE THIS
\usepackage{natbib}  % DO NOT CHANGE THIS AND DO NOT ADD ANY OPTIONS TO IT
\usepackage{caption} % DO NOT CHANGE THIS AND DO NOT ADD ANY OPTIONS TO IT
\usepackage{algorithm}
\usepackage{bm}

\usepackage{newfloat}
\usepackage{listings}
\DeclareCaptionStyle{ruled}{labelfont=normalfont,labelsep=colon,strut=off} % DO NOT CHANGE THIS
\floatstyle{ruled}
\newfloat{listing}{tb}{lst}{}
\floatname{listing}{Listing}
\usepackage{array} % Needed for fixed width columns
\newcolumntype{P}[1]{>{\centering\arraybackslash}p{#1}}

\usepackage{tcolorbox}
\usepackage{booktabs}
\usepackage{tabularx}
\usepackage{multirow}
\usepackage{amssymb}
\usepackage{times}
\usepackage{soul}
\usepackage{url}
\usepackage[utf8]{inputenc}
\usepackage{amsmath}
\usepackage{amsthm}
\usepackage{booktabs}
\usepackage[switch]{lineno}
\usepackage{algorithm}
\usepackage{algpseudocode}

\title{Empirical Analysis of Privacy-Fairness-Accuracy Trade-offs in Federated Learning: A Step Towards Responsible AI}

\author {
    Dawood Wasif\textsuperscript{\rm 1},
    Dian Chen\textsuperscript{\rm 1},
    Sindhuja Madabushi\textsuperscript{\rm 1},
    Nithin Alluru\textsuperscript{\rm 1},
    Terrence J Moore\textsuperscript{\rm 2},
    Jin-Hee Cho\textsuperscript{\rm 1}
}

\affiliations {
    \textsuperscript{\rm 1}Virginia Tech, Blacksburg, Virginia, United States\\
    \textsuperscript{\rm 2}US Army Research Laboratory, Adelphi, Maryland, United States\\
    \{dawoodwasif, dianc, msindhuja, nithin, jicho\}@vt.edu, terrence.j.moore.civ@army.mil
}

\usepackage{bibentry}
\begin{document}

\maketitle

\begin{abstract}
Federated Learning (FL) enables collaborative model training while preserving data privacy; however, balancing privacy preservation (PP) and fairness poses significant challenges. In this paper, we present the first unified large-scale empirical study of privacy--fairness--utility trade-offs in FL, advancing toward responsible AI deployment. Specifically, we systematically compare Differential Privacy (DP), Homomorphic Encryption (HE), and Secure Multi-Party Computation (SMC) with fairness-aware optimizers including q-FedAvg, q-MAML, Ditto, evaluating their performance under IID and non-IID scenarios using benchmark (MNIST, Fashion-MNIST) and real-world datasets (Alzheimer’s MRI, credit-card fraud detection). Our analysis reveals HE and SMC significantly outperform DP in achieving equitable outcomes under data skew, although at higher computational costs. Remarkably, we uncover unexpected interactions: DP mechanisms can negatively impact fairness, and fairness-aware optimizers can inadvertently reduce privacy effectiveness. We conclude with practical guidelines for designing robust FL systems that deliver equitable, privacy-preserving, and accurate outcomes.
\end{abstract}

% Uncomment the following to link to your code, datasets, an extended version or similar.
\begin{links}
    \link{Code}{https://github.com/dawoodwasif/Privacy_vs_Fairness_vs_Accuracy}
    \link{Extended version}{https://arxiv.org/abs/2503.16233}
\end{links}

\section{Introduction} \label{sec:introduction}

Federated Learning (FL) enables decentralized model training while preserving privacy, yet its deployment in realistic scenarios such as healthcare and finance raises challenges at the intersection of privacy, fairness, and real-world impact. Privacy-preserving techniques like Differential Privacy (DP), Homomorphic Encryption (HE), and Secure Multi-Party Computation (SMC) enhance security but introduce fairness concerns. DP’s noise protects data but degrades utility for underrepresented groups, exacerbating bias~\cite{bagdasaryan2019differential}. HE and SMC preserve data integrity and mitigate fairness disparities but impose additional computational costs, potentially limiting participation~\cite{truex2019hybrid, xu2023industry}. Despite advancements, privacy-fairness trade-offs in FL remain underexplored, necessitating privacy-aware fairness mechanisms that align with responsible AI to ensure equitable and transparent model performance.

A critical but often overlooked issue is {\em client fairness}, ensuring equitable model performance across clients with heterogeneous data and resources. Unlike algorithmic fairness, which addresses bias in decisions, client fairness examines FL’s impact on diverse clients. In dermatology, FL models trained on imbalanced client data favored overrepresented skin types, reducing accuracy for underrepresented groups~\cite{weng2020addressing}, highlighting the need for bias mitigation in healthcare AI.

To promote AI for social good, privacy-preserving mechanisms must integrate fairness by collaborating with stakeholders, including healthcare institutions and policymakers. Addressing computational and data heterogeneity while maintaining equitable performance is key to preventing societal inequalities. Our research explores privacy-fairness trade-offs in FL, developing frameworks that align with global fairness initiatives like the United Nations Sustainable Development Goals (SDGs) and the Leave No One Behind (LNOB) Principle~\cite{bentaleb2024survey}.

% Prior research has largely treated privacy and fairness in federated learning as separate concerns and explored each within narrow settings. Privacy-focused studies have evaluated differential privacy and encryption schemes under idealized IID assumptions without accounting for fairness metrics or realistic adversarial threats~\cite{truex2020ldp, wang2020global, zhang2022homomorphic, liu2022privacy}. Conversely, fairness-aware approaches typically assess equity under controlled conditions but omit privacy-preserving mechanisms and systematic attack analyses~\cite{li2019fair, mohri2019agnostic}. Moreover, most empirical work relies on single benchmark datasets or small-scale simulations and rarely validates methods on datasets with real-world applicability. Such fragmented efforts leave open the question of how DP, HE, and SMC interact with fairness optimizers across heterogeneous client distributions and practical domains.

Prior work in federated learning often treats privacy and fairness separately, with privacy studies focusing on differential privacy and encryption under idealized IID settings while ignoring fairness metrics and realistic adversarial threats~\cite{truex2020ldp, wang2020global, zhang2022homomorphic, liu2022privacy}. Conversely, fairness-aware approaches assess equity under controlled conditions but omit privacy-preserving mechanisms and attack analyses~\cite{li2019fair, mohri2019agnostic}. Most evaluations also rely on limited benchmarks or small-scale simulations, leaving open how DP, HE, and SMC interact with fairness optimizers in heterogeneous and practical domains.

{\bf The goal of this work} is to investigate the complex interactions between privacy preservation and fairness in federated learning across both benchmark and real-world scenarios in healthcare and finance. We assess the impact of Differential Privacy (DP), Homomorphic Encryption (HE), and Secure Multi-Party Computation (SMC) on client-level fairness metrics and evaluate whether fairness-aware optimization strategies introduce new privacy vulnerabilities. Our empirical study uses four representative datasets: MNIST and Fashion-MNIST for controlled benchmarks, Alzheimer’s Disease MRI scans for medical applications, and a large-scale credit card fraud detection dataset for financial analysis. By conducting experiments under both IID and non-IID client distributions, we aim to chart the multi-dimensional trade-offs that practitioners face when deploying FL in sensitive domains.

By quantifying the trade-offs between privacy and fairness, we deliver actionable guidance for designing federated learning frameworks that uphold strong confidentiality guarantees while ensuring equitable outcomes across diverse clients. Our results offer domain-specific recommendations under regulations such as HIPAA and GDPR for medical use cases and the EU AI Act for financial services. We further propose a set of best practices and parameter guidelines that practitioners can adopt to balance differential privacy budgets, encryption parameters, and fairness weights. This work advances responsible AI by providing empirical evidence and practical strategies that mitigate fairness disparities introduced by privacy mechanisms, enabling trustworthy FL deployment in real-world settings.

\noindent
This work presents the following {\bf key contributions}:
\begin{itemize}
  \item We present the first unified, large-scale empirical evaluation of DP, HE, and SMC under fairness-aware optimizers on both IID and non-IID client distributions across benchmark and real-world datasets.
\item We systematically quantify trade-offs between privacy-preserving techniques and client fairness under simulated attacks, offering detailed insights into their behavior in heterogeneous federated learning.

  \item Our analysis shows that HE and SMC achieve a superior balance between privacy and fairness compared to DP, particularly in non-IID scenarios where DP noise disproportionately degrades utility for underrepresented clients, while HE incurs higher computational overhead.
  \item We uncover counterintuitive interactions between privacy mechanisms and fairness optimizers, demonstrating that DP can erode fairness and that fairness-aware updates can weaken privacy defenses.
 \item We propose a comprehensive validation framework integrating privacy-fairness schemes with adversarial threat models, offering practical guidelines for robust federated learning in healthcare and finance applications.

\end{itemize}

\section{Related Work} \label{sec:related-work}

\subsubsection{Privacy-Preserving FL} Various studies propose privacy-preserving (PP) techniques for FL, often combining DP and SMC to balance accuracy and privacy, as SMC risks inference attacks and DP reduces accuracy~\cite{truex2019hybrid}. Methods like Confined Gradient Descent and blockchain-based optimization reduce cryptographic overhead~\cite{zhang2024privacypreserving}. Cryptographic tools, Trusted Execution Environments (TEEs), and decentralized frameworks are critical, especially in healthcare FL. HE defends against reconstruction attacks~\cite{zhang2022homomorphic}, while TEE-based FL secures clients and servers~\cite{yazdinejad2023ap2fl}. Decentralized FL enhances robustness and communication efficiency~\cite{tian2023robust}. PP FL models support Alzheimer’s and COVID-19 detection by safeguarding sensitive data~\cite{li2022health}.

Despite advances, fairness impacts remain underexplored, and standardized privacy metrics are lacking. Theoretical frameworks dominate, with limited real-world benchmarking, restricting FL applicability.

\subsubsection{Fairness-Aware FL} Existing studies address fairness in FL by ensuring balanced model performance across heterogeneous client data. Most focus on group fairness, aiming for equitable outcomes across demographic groups~\cite{mohri2019agnostic}, but overlook client fairness (see Supplementary Material, Section B), which considers variations in dataset sizes, distributions, and resources. Other approaches emphasize fairness in client contributions and performance using aggregation techniques and constrained optimization, particularly in healthcare~\cite{meerza2022fair} and finance~\cite{kamalaruban2024evaluating}. However, these primarily maintain fairness at the group level rather than ensuring equitable benefits for clients.

Despite progress, fairness-aware methods often lack privacy-preserving mechanisms, leaving FL systems vulnerable to privacy risks. Moreover, many assume IID data, while real-world federated settings are predominantly non-IID~\cite{zhao2018federated}, highlighting the need for rigorous fairness evaluations under privacy constraints.

\subsubsection{Critical Tradeoffs in FL}  Balancing privacy and fairness in FL remains challenging. Privacy-preserving techniques like DP and HE reduce accuracy, disproportionately affecting underrepresented groups~\cite{Chen23-csur-survey}. Conversely, fairness-enhancing methods, such as subgroup-focused optimization, may heighten privacy risks by overfitting minority groups~\cite{zheng2023trojfair}.  To mitigate these issues, two-stage frameworks enforce fairness (e.g., demographic parity) before applying privacy techniques like local DP~\cite{padala2021federated}. Integrated approaches~\cite{corbucci2024puffle,pentyala2022privfairfl} optimize privacy and fairness during training. 

Hence, unlike prior work that treats privacy and fairness separately, our framework jointly quantifies the privacy–fairness–utility trade-off under both IID and non-IID settings.

% \subsubsection{Addressed Key Research Gaps} Existing work lacks a unified framework to quantify privacy-fairness-utility trade-offs in FL under both IID and non-IID data distributions. It often treats privacy and fairness separately, limiting joint analysis. This gap is critical in high-risk domains like healthcare and finance. Our study aims to ensure client fairness, ensuring equitable benefits for all FL participants rather than just group fairness. 

\section{Experimental Validation Framework}
\label{sec:exp-setup}

\subsection{Datasets}
To systematically explore the interplay between privacy, fairness, and accuracy in FL, we employ four datasets spanning benchmark tasks, medical imaging, and financial fraud detection. Each dataset is partitioned under both IID and non‐IID regimes to capture realistic client heterogeneity.  

\begin{itemize}
  \item \textbf{MNIST:} The MNIST dataset~\cite{lecun1998mnist} consists of 70,000 grayscale images (28$\times$28 resolution) of handwritten digits. Data is normalized to [0,1] and partitioned across $K=50$ clients. IID partitions distribute data uniformly, while non-IID partitions follow a Dirichlet-based approach~\cite{li2022federated} ($\alpha=0.5$) to simulate realistic data heterogeneity.

  \item \textbf{Fashion-MNIST:} Fashion-MNIST~\cite{xiao2017fashion} includes 70,000 grayscale images (28$\times$28 resolution) across 10 clothing categories. Preprocessing and partitioning methods are identical to MNIST, with the same Dirichlet parameter ($\alpha=0.5$), introducing increased visual complexity to test robustness.

  \item \textbf{Alzheimer’s Disease MRI (AD-MRI):} The AD-MRI dataset~\cite{alzheimer_mri_dataset} comprises 4,320 T1-weighted MRI scans labeled Alzheimer’s or healthy. After preprocessing (skull-stripping, normalization, resizing to 64$\times$64), SMOTE~\cite{chawla2002smote} balances classes. Data is distributed across $K=50$ hospitals, with IID splits equally balanced, and non-IID splits reflecting specialized (80\% Alzheimer’s) versus general hospitals (80\% healthy).

  \item \textbf{Credit Card Fraud Detection (CCFD):} The CCFD dataset~\cite{dal2017credit} contains 284,807 transactions (0.17\% fraudulent). After standardization, SMOTE oversamples fraud cases to 5\%. Data is distributed among $K=50$ institutions: IID splits maintain equal fraud rates, whereas non-IID splits vary fraud proportions (10\%, 1\%, and 0.5\%), simulating realistic banking heterogeneity.
\end{itemize}

\subsection{Hyperparameters}
Hyperparameters were tuned via grid search on a baseline FL model without privacy or fairness enhancements and applied across all FL variants. The key parameters include a learning rate of \textit{0.1}, batch size of \textit{512}, \textit{40} local epochs per client, \textit{20} communication rounds, and \textit{5} participating clients per round. Table~\ref{tab:param-default} summarizes the default values. A CNN with two convolutional layers (32, 64 filters), max-pooling, and a dense layer (1024 units) was used for MNIST, Fashion-MNIST, and MRI, whereas Logistic Regression with L2 regularization (0.01) was used for CCFD.

\begin{table}[t]
\centering
\caption{\sc Key Design Parameters and Their Default Values}
\label{tab:param-default}
% \vspace{-2mm}
\begin{tabular}{|P{4.5cm}|P{1.5cm}|}
\hline
\textbf{Parameter} & \textbf{Value} \\
\hline
Batch size & 512 \\
\hline
Communication rounds & 20 \\
\hline
Local training epochs & 40 \\
\hline
Simulation runs & 10 \\
\hline
Learning rate & 0.1 \\
\hline
Learning rate lambda & 0.1 \\
\hline
Total number of clients & 50 \\
\hline
Fraction of clients per round & 0.1 \\
\hline
\end{tabular}
% \vspace{-2mm}
\end{table}

\subsection{Threat Model}
\label{sec:threat-model}

We assume a single, adaptive adversary $\mathcal{A}$ capable of compromising up to $\beta K$ clients, where $K$ is the total number of clients and $\beta \in [0,1]$ is the maximum fraction under adversarial control. In a white‐box scenario, $\mathcal{A}$ observes all model updates, gradient exchanges, and protocol metadata but never has direct access to raw client data. This threat model covers two categories of attacks: privacy attacks, which aim to extract sensitive information from model updates, and fairness attacks, which introduce or exacerbate performance disparities among clients.

\subsubsection*{Privacy Attacks}

\paragraph{Membership Inference Attacks}  
These attacks aim to determine whether a specific record was used in training by exploiting model memorization. The adversary collects global model updates and client gradients over multiple rounds, then trains shadow models to approximate client data distributions. By comparing the target model’s confidence scores or losses on candidate samples with those of the shadow models, $\mathcal{A}$ can infer membership. The \textit{Membership Inference Attack Success Ratio (MSR)} measures privacy leakage as the fraction of correct guesses minus false positives.

\paragraph{Differential Leakage Attacks}  
These attacks assess how aggregated updates change with the inclusion or removal of a single “canary” record. The adversary alternates between injecting and removing the record locally and observes differences in aggregated gradients. Averaging the $\ell_1$‐norm differences per parameter across rounds yields the \textit{Differential Leakage Rate (DLR)}, where higher values indicate greater detectability of minor client contributions. DLR complements MSR by focusing on gradient sensitivity rather than model outputs.

% \subsubsection*{Privacy Attacks}

% \paragraph{Membership Inference Attacks}  
% Membership inference attacks attempt to determine whether a specific data record was used during model training by exploiting memorization effects in model parameters. The adversary first gathers global model updates and individual client gradients across multiple rounds. Using these observations, shadow models are trained to approximate client data distributions. By comparing the target model’s confidence scores or loss values on candidate samples against those of the shadow models, $\mathcal{A}$ can infer membership. The \textit{Membership Inference Attack Success Ratio (MSR)} quantifies this threat as the fraction of correct membership guesses minus false positives, providing a direct measure of privacy leakage through model outputs and gradients.

% \paragraph{Differential Leakage Attacks}  
% Differential leakage attacks probe how sensitive aggregated updates are to the inclusion or exclusion of single records. The adversary injects or removes a “canary” record in its local dataset and observes the resulting change in the aggregated gradient. By repeating this procedure over many training rounds, small update differences are amplified. The \textit{Differential Leakage Rate (DLR)} is defined as the average $\ell_1$‐norm difference per parameter between updates with and without the canary. A higher DLR indicates weaker perturbation‐based defenses, as minor changes in client contributions become detectable. DLR complements MSR by focusing on gradient sensitivity rather than output behavior.

\subsubsection*{Fairness Attacks}

\paragraph{Data Poisoning Attacks}  
In data poisoning attacks, the adversary injects malicious samples into 10\% of compromised clients’ training data (by changing labels randomly) to bias the global model. We quantify impact using \textit{Data Poisoning Attack Accuracy (DPA-A)}, the reduction in clean test accuracy, and \textit{Data Poisoning Attack Accuracy Disparity (DPA-AD)}, the variance in accuracy across clients. High DPA-AD indicates uneven harm, revealing fairness weaknesses.

\paragraph{Backdoor Attacks}  
Backdoor attacks plant triggers in 10\% of training samples so that inputs bearing the trigger are misclassified to a target label while clean-data accuracy remains stable. We measure this with \textit{Backdoor Attack Accuracy (BA-A)}, the success rate on triggered inputs, and \textit{Backdoor Attack Accuracy Disparity (BA-AD)}, the client-level variance in BA-A. Elevated BA-AD highlights uneven vulnerability to backdoors across participants.

% \textbf{Lattice Attacks on Homomorphic Encryption.}  
% When federated gradients are protected by homomorphic encryption (HE), the adversary loses direct access to plaintext updates but may still attempt a lattice‐reduction attack to recover the secret key.  By collecting multiple ciphertexts produced under the same encryption parameters, $\mathcal{A}$ constructs an instance of the Learning With Errors (LWE) problem and applies lattice‐basis reduction algorithms.  Successful lattice attacks break the cryptographic layer and expose all subsequent gradients in plaintext.  This threat underscores the importance of selecting HE parameters (modulus sizes, noise distributions, polynomial degrees) that resist known lattice reduction strategies.

% \textbf{Share Reconstruction Attacks in Secure Multi‐Party Computation.}  
% In SMC‐based federated learning, each client’s gradient is split into multiple shares and distributed among different parties.  The reconstruction threshold is set so that only an authorized subset of shares can recover the original gradient.  An adversary that compromises enough clients can collude to pool their shares and perform Lagrange interpolation, reconstructing the private gradient.  This attack highlights the tension between fault tolerance (lower thresholds improve robustness to dropouts) and privacy (higher thresholds resist collusion), and emphasizes the need to carefully configure share counts and thresholds in SMC protocols.

\subsection{Metrics}
\label{sec:metrics}

To compare privacy‐preserving and fairness‐aware FL methods under benign and adversarial conditions, we employ a broad set of quantitative metrics grouped into privacy, fairness, and utility categories. Each metric is computed at the end of training (round $T$) and averaged over ten independent runs. We report both the mean and the standard deviation.

\subsubsection{Privacy Metrics} 
These metrics serve as practical indicators of a federated learning system’s vulnerability by translating privacy guarantees into observable outcomes under realistic adversarial conditions. Rather than relying exclusively on theoretical privacy budgets or cryptographic assumptions, we adopt an empirical perspective that directly measures information leakage through attack simulations. Consequently, our analysis centers on \textit{Membership Inference Attack Success Ratio} (MSR) and \textit{Differential Leakage Rate} (DLR).

\begin{itemize}
  \item \textbf{Membership Inference Attack Success Ratio (MSR):}  
  This metric captures the adversary’s ability to distinguish training samples from non-training ones. While various attack strategies exist, we abstract the outcome into a scalar by comparing the model’s confidence on held-in versus held-out samples. Let $p_w(y \mid x)$ denote the model’s predicted probability for the true label $y$ given input $x$. We define:
  \begin{equation}
  \textbf{MSR} = 
  \frac{1}{2N} \sum_{i=1}^{N} \left[ p_w(y_i \mid x_i^{\mathrm{in}}) - p_w(y_i \mid x_i^{\mathrm{out}}) \right],
  \label{eq:msr}
  \end{equation}
  where $\{(x_i^{\mathrm{in}}, y_i)\}$ are samples from the training set and $\{(x_i^{\mathrm{out}}, y_i)\}$ from an equally sized held-out set. A higher MSR indicates greater separation between training and non-training confidences, implying weaker privacy.

  \item \textbf{Differential Leakage Rate (DLR):}  
  This metric evaluates leakage under perturbation-based defenses (e.g., differential privacy) by quantifying how gradients change when a single data point is added or removed. Let $\nabla F_k^{+}(w)$ and $\nabla F_k^{-}(w)$ be the gradients from client $k$ with and without the target example. We define:
  \begin{equation}
  \textbf{DLR} = 
  \frac{1}{d} \left\| \nabla F_k^{+}(w^{(T)}) - \nabla F_k^{-}(w^{(T)}) \right\|_{1},
  \label{eq:dlr}
  \end{equation}
  where $d$ is the total number of model parameters. A lower DLR indicates stronger protection against differential leakage.
\end{itemize}
% \smallskip
% \textbf{Lattice Attack Resistance (LSR).}  
% For homomorphic encryption schemes, we assess the difficulty of recovering the secret key via lattice reduction.  After collecting encrypted gradients, an attacker constructs an LWE lattice and performs basis reduction.  We encapsulate this resistance by measuring the residual error in the recovered key.  If $s$ is the true secret and $\hat s$ the attacker’s estimate, then
% \[
% \textbf{LSR} \;=\; 
% 1 \;-\; \frac{\|\hat s - s\|_{2}}{\|s\|_{2}}.
% \]
% Values close to one indicate that the recovered key is almost as large in error as the true key itself, implying high resistance to lattice attacks.

% \smallskip
% \textbf{Share Reconstruction Robustness (SAR).}  
% In secure multi‐party computation, gradients are split into $N_s$ shares and require a threshold $t$ for reconstruction.  We evaluate how accurately an adversary who collects $t$ shares can rebuild the original gradient.  Denote $\nabla F_k$ the true gradient and $\hat{\nabla F}_k$ its reconstruction; we define
% \[
% \textbf{SAR} \;=\; 
% 1 \;-\; \frac{\|\hat{\nabla F}_k - \nabla F_k\|_{2}}{\|\nabla F_k\|_{2}}.
% \]
% A higher SAR indicates that reconstruction yields a vector almost orthogonal to the true gradient, signifying strong resilience to collusion.

\subsubsection{Fairness Metrics}
These metrics evaluate how evenly model performance is distributed across participating clients, ensuring that privacy or optimization choices do not disproportionately harm any subset of users.
\begin{itemize}
  \item \textbf{Loss Disparity (LD):}  
  To assess how equitably the global model performs across clients, we evaluate the variance of per-client training losses. Let $\mathcal{L}_k$ be the empirical loss on client $k$'s local data under the global model $w^{(T)}$. We define:
  \begin{equation}
  \textbf{LD} = 
  \frac{1}{K} \sum_{k=1}^{K} \left( \mathcal{L}_k - \overline{\mathcal{L}} \right)^2, 
  \quad 
  \overline{\mathcal{L}} = \frac{1}{K} \sum_{k=1}^{K} \mathcal{L}_k,
  \label{eq:ld}
  \end{equation}
  where lower LD values indicate more uniform training loss distribution, promoting fairness.

  \item \textbf{Accuracy Disparity (AD):}  
  This metric captures the variance in client-level model accuracy. Let $\mathrm{Acc}_k$ denote the accuracy on client $k$’s held-out data. We define:
  \begin{equation}
  \textbf{AD} = 
  \frac{1}{K} \sum_{k=1}^{K} \left( \mathrm{Acc}_k - \overline{\mathrm{Acc}} \right)^2, 
  \quad 
  \overline{\mathrm{Acc}} = \frac{1}{K} \sum_{k=1}^{K} \mathrm{Acc}_k,
  \label{eq:ad}
  \end{equation}
  where lower AD (also referred to as local variance, LV) reflects more consistent model benefit across clients, indicating stronger fairness.
\end{itemize}

\subsubsection{Utility Metrics}
The utility metrics ensure that privacy or fairness enhancements do not unduly degrade the global model's overall predictive performance.

\begin{itemize}
  \item \textbf{Global Accuracy (Acc):}  
  We assess overall model performance on a held-out test set of size $N$. Let $\hat{y}_i$ be the predicted label and $y_i$ the true label for example $i$. We define:
  \begin{equation}
  \textbf{Acc} = 
  \frac{1}{N} \sum_{i=1}^{N} \mathbf{1}(\hat{y}_i = y_i),
  \label{eq:acc}
  \end{equation}
  where $\mathbf{1}(\cdot)$ is the indicator function. This standard metric verifies whether privacy and fairness defenses preserve overall classification quality.
\end{itemize}

\subsection{Federated Learning Schemes}
We examine both Fair and Privacy-Preserving (PP) FL schemes to analyze fairness-privacy-accuracy tradeoffs.  
\subsubsection{Fair FL Schemes} \begin{itemize}
  \item \texttt{q-FedAvg}~\cite{li2019fair}: Extends FedAvg by giving more weight to clients with higher losses. Fairness is controlled by the parameter \(q\), where \(q = 0\) reduces the method to standard FedAvg.

  \item \texttt{q-FedSGD}~\cite{li2019fair}: Builds on q-FedAvg using FedSGD, where clients send gradient updates after each iteration. The server aggregates them using a \(q\)-weighted objective, prioritizing high-loss clients.

  \item \texttt{q-MAML}~\cite{li2019fair}: Incorporates q-based fairness objectives into Model-Agnostic Meta-Learning (MAML)~\cite{finn2017model}, adjusting global updates using a \(q\)-weighted loss function.

  \item Agnostic Federated Learning (\texttt{AFL})~\cite{mohri2019agnostic}: Uses a minimax optimization framework to improve worst-case performance by minimizing the maximum loss over possible data distributions.

  \item \texttt{Ditto}~\cite{li2021ditto}: Simultaneously trains a global model and a personalized model for each client, enabling personalized performance while maintaining generalization across clients.
\end{itemize}

\subsubsection{Privacy-Preserving (PP) FL Schemes} \begin{itemize}
  \item \textbf{\em Perturbation-based PP FL:}  
  \texttt{Local Differential Privacy (LDP)}~\cite{truex2020ldp} adds noise at the client level to protect individual data, though it may reduce model accuracy.  
  \texttt{Global Differential Privacy (GDP)}~\cite{wang2020global} injects noise at the server level to mask aggregated updates, offering group-level privacy at the cost of higher noise.  
  \texttt{Gradient Masking (GM)}~\cite{boenisch2021gradient} obfuscates gradients via noise and clipping, similar to LDP, to defend against adversarial reconstruction.

  \item \textbf{\em Anonymization-based PP FL:}  
  \texttt{$K$-Anonymity} ensures each data record is indistinguishable from at least \( k-1 \) others by grouping gradient updates~\cite{sweeney2002achieving}.  
  \texttt{$L$-Diversity}~\cite{machanavajjhala2007diversity} enhances $K$-anonymity by requiring at least \( l \) distinct sensitive values per group.  
  \texttt{$T$-Closeness}~\cite{li2006t} ensures that the distribution of sensitive attributes within any group closely matches the global distribution, minimizing data pattern leakage.

  \item \textbf{\em Encryption-based PP FL:}  
  \texttt{HE}~\cite{fang2021privacy} allows computations on encrypted data, keeping gradients hidden throughout training.  
  \texttt{SMC}~\cite{liu2022privacy} enables joint computation across parties without revealing individual data, maintaining confidentiality.

  \item \textbf{\em Hybrid PP FL Techniques:} \texttt{LDP+HE} combines client-side noise and encrypted computation to balance individual privacy and security.  \texttt{GDP+HE} applies server-side noise while securing updates via encryption.  \texttt{LDP+SMC} integrates client-level noise with secure collaborative computation.  \texttt{GDP+SMC} ensures group-level privacy via cryptographic protection.  \texttt{HE+SMC} fuses encryption and secure computation for end-to-end PP training.
\end{itemize}

For the joint analysis of privacy-fairness-accuracy tradeoffs, we developed an FL framework integrating \texttt{q-FedAvg} with LDP, GDP, HE, and SMC, as detailed in Algorithm~\ref{qfedavg}.  

\begin{algorithm}[t]
%\small 
\caption{\texttt{q-FedAvg} with LDP, GDP, HE, and SMC}\label{qfedavg}
\begin{algorithmic}[1]
\State \textbf{Input:} $K$, $E$, $T$, $q$, $1/L$, $\eta$, $w^0$, $p_k$, $k = 1, \dots, m$
\State \textbf{Initialize:} Global model $w^{(0)}$
\For{$t = 0, \dots, T-1$}
    \State Server selects a subset $S_t$ of $K$ devices at random (each device $k$ is chosen with prob. $p_k$)
    \State Server sends $w^t$ to all selected devices
    \For{each selected device $k \in S_t$ \textbf{in parallel}}
        \State Each selected device $k$ updates $w^t$ for $E$ epochs of SGD on $F_k$ with step-size $\eta$ to obtain $\bar{w}_k^{t+1}$
        \State Compute local weight update $\Delta w_k^t = L(w^t - \bar{w}_k^{t+1})$
        \State Compute fairness-adjusted gradient $\Delta_k^t = F_k^q(w^t)\Delta w_k^t$
        \State Calculate weight $h_k^t = qF_k^{q-1}(w^t)\|\Delta w_k^t\|^2 + LF_k^q(w^t)$
        \State \textbf{Apply LDP or GDP:}
        \State \hspace{1em} \textbf{LDP:} $\tilde{\Delta}_k^t = \Delta_k^t + \mathcal{N}(0, \Delta^2/\epsilon^2)$
        \State \hspace{1em} \textbf{GDP:} $\tilde{\Delta}_k^t = \sum_{k} \Delta_k^t + \mathcal{N}(0, \Delta^2/\epsilon^2)$ 
        
        \State \textbf{Apply HE:} $\mathcal{E}_k(\tilde{\Delta}_k^t) = \text{Enc}(\tilde{\Delta}_k^t)$
        \State \textbf{Apply SMC:} Split $\tilde{\Delta}_k^t$ into shares $\tilde{\Delta}_k^{tj}$ 
    \EndFor
    \State Each selected device $k$ sends $\tilde{\Delta}_k^t$ and $h_k^t$ back to the server
    \State Server aggregates updates as $w^{t+1} = w^t - \frac{\sum_{k \in S_t} \tilde{\Delta}_k^t}{\sum_{k \in S_t} h_k^t}$
\EndFor
\State \textbf{Return} final model $w^{(T)}$
\end{algorithmic}
\end{algorithm}
% The source code and experimental framework are publicly available on GitHub\footnote{\url{https://github.com/dawoodwasif/Privacy_vs_Fairness_vs_Accuracy}}.

\section{Simulation Results \& Analysis} \label{sec:results-analyses}

%This section evaluates the proposed framework under varying fairness and privacy conditions across three datasets in Section~\nameref{sec:exp-setup}. A detailed analysis of the results follows.
% \begin{comment}
% \begin{figure*}[ht]
%     \centering
%     \includegraphics[width=\textwidth]{figures/graph3-latest.png} 
%     \caption{Impact of privacy-preserving FL algorithms (GDP, LDP, HE, and SMC) across four datasets under IID and non-IID conditions, evaluated for different \( q \)-values (0, 1, 5, 10).}
%     \label{fig:graph1} % Replace with your label for referencing
% \end{figure*}
% \end{comment}

\subsection{Effect of Varying Fairness on Privacy Risk}
\label{subsec:fairness-privacy-risk}

Table~\ref{tab:iid_q_sensitive} reports the most striking privacy‐risk variations when adjusting the fairness weight \(q\) under IID splits. On MNIST with global DP, loss disparity (LD) climbs from 0.0227 at \(q=0\) to 0.0837 at \(q=10\), differential attack leakage rate (DLR) rises from 0.00267 to 0.00443, and membership‐inference attack success ratio (MSR) oscillates near 0.40. Local DP on Fashion‐MNIST shows a dramatic drop in LD from 0.447 to 0.00013, while DLR shifts modestly (0.00777→0.00877) and MSR remains in the 0.48–0.51 range. Homomorphic encryption on Fashion‐MNIST follows a similar pattern (LD 0.166→0.00004, DLR 0.00220→0.00430, MSR 0.59→0.63). Secure MPC on MNIST yields a reduction in LD (0.0220→0.00483) with only small increases in DLR (0.00059→0.00239) and MSR (0.59→0.60). Under IID MRI, only local DP is sensitive (LD up to 0.00059, DLR to 0.00880, MSR to 0.59), and on IID CCFD all schemes except DP show negligible changes.

\begin{table*}[t]
\centering
\caption{$q$‐value sensitivity under IID settings: Only schemes and metrics exhibiting significant variation are shown.}
\label{tab:iid_q_sensitive}
\footnotesize
\renewcommand{\arraystretch}{1.1}
\begin{tabular}{l l r r r  l l r r r}
\toprule
\multicolumn{5}{c}{\textbf{MNIST (IID)}} & \multicolumn{5}{c}{\textbf{FMNIST (IID)}} \\
\cmidrule(lr){1-5} \cmidrule(lr){6-10}
Scheme & Metric & $q=0$ & $q=1$ & $q=10$ 
       & Scheme & Metric & $q=0$ & $q=1$ & $q=10$ \\
\midrule
GDP  & LD  & 0.0227 & 0.0490 & 0.0837 
     & GDP  & ACC & 0.333  & 0.356  & 0.316  \\
     & DLR  & 0.00267 & 0.00449 & 0.00443 
     &      & LD  & 27.25  & 11.50  &  5.46  \\
     & MSR & 0.390   & 0.350   & 0.410   
     &      & DLR  & 0.00279 & 0.00439 & 0.00442 \\
HE   & LD  & 0.0288 & 0.0168 & 0.00481 
     & LDP  & LD  & 0.447   & 0.00220 & 0.00013 \\
     & DLR  & 0.00225 & 0.00261 & 0.00428 
     &      & DLR  & 0.00777 & 0.00825 & 0.00877 \\
     & MSR & 0.560   & 0.520   & 0.530   
     &      & MSR & 0.480   & 0.510   & 0.470   \\
SMC  & LD  & 0.0220 & 0.0165 & 0.00483 
     & HE   & LD  & 0.166   & 0.00496 & 0.00004 \\
     & DLR  & 0.00059 & 0.00087 & 0.00239 
     &      & DLR  & 0.00220 & 0.00252 & 0.00430 \\
     & MSR & 0.590   & 0.600   & 0.600   
     &      & MSR & 0.590   & 0.660   & 0.630   \\
\midrule
\multicolumn{5}{c}{\textbf{MRI (IID)}} & \multicolumn{5}{c}{\textbf{CCFD (IID)}} \\
\cmidrule(lr){1-5} \cmidrule(lr){6-10}
Scheme & Metric & $q=0$  & $q=1$   & $q=10$ 
       & Scheme & Metric & $q=0$    & $q=1$    & $q=10$     \\
\midrule
LDP  & LD  & 0.00012 & 0.00044 & 0.00059 
     & LDP  & LD  & 0.00001  & 0.00004  & 0.000001  \\
     & DLR  & 0.00742 & 0.00842 & 0.00880 
     &      & DLR  & 0.00757  & 0.00842  & 0.00881   \\
     & MSR & 0.520   & 0.590   & 0.550   
     &      & MSR & 0.430    & 0.470    & 0.480     \\
HE   & LD  & 0.00064 & 0.00015 & 0.00074 
     & HE   & LD  & 0.00001  & 0.00004  & 0.000001  \\
     & DLR  & 0.00222 & 0.00256 & 0.00428 
     &      & DLR  & 0.00225  & 0.00261  & 0.00428   \\
SMC  & LD  & 0.00086 & 0.00056 & 0.00039 
     & SMC  & LD  & 0.00001  & 0.00004  & 0.000001  \\
     & DLR  & 0.00058 & 0.00086 & 0.00249 
     &      & DLR  & 0.00059  & 0.00087  & 0.00239   \\
\bottomrule
\end{tabular}
\end{table*}

\begin{table*}[h!]
\centering
\caption{$q$‐value sensitivity under Non‐IID settings: Only schemes and metrics exhibiting significant variation are shown.}
\label{tab:non_iid_q_sensitive}
\footnotesize
\renewcommand{\arraystretch}{1.1}
\begin{tabular}{l l r r r  l l r r r}
\toprule
\multicolumn{5}{c}{\textbf{MNIST (Non‐IID)}} & \multicolumn{5}{c}{\textbf{FMNIST (Non‐IID)}} \\
\cmidrule(lr){1-5} \cmidrule(lr){6-10}
Scheme & Metric & $q=0$ & $q=1$ & $q=10$ 
       & Scheme & Metric & $q=0$ & $q=1$ & $q=10$ \\
\midrule
GDP  & LD  & 0.0627 & 0.1640 & 0.0926 
     & GDP  & ACC & 0.283  & 0.617  & 0.352  \\
     & DLR  & 0.00567 & 0.00551 & 0.00533
     &      & LD  & 0.443  & 0.00866 & 0.0997 \\
     & MSR & 0.550   & 0.480   & 0.550  
     &      & DLR  & 0.00357 & 0.00573 & 0.00581 \\
HE   & LD  & 0.0794 & 0.0911 & 0.0028
     & LDP  & LD  & 0.0338 & 0.0106 & 0.00392 \\
     & DLR  & 0.00328 & 0.00399 & 0.00512
     &      & DLR  & 0.00891 & 0.00904 & 0.00936 \\
     & MSR & 0.700   & 0.760   & 0.750  
     &      & MSR & 0.600   & 0.650   & 0.710   \\
SMC  & LD  & 0.0812 & 0.0909 & 0.00973
     & HE   & LD  & 0.596  & 0.00341 & 0.00269 \\
     & DLR  & 0.00812 & 0.00095 & 0.00274
     &      & DLR  & 0.00220 & 0.00252 & 0.00430 \\
     & MSR & 0.750   & 0.770   & 0.750  
     &      & MSR & 0.590   & 0.660   & 0.630   \\
\midrule
\multicolumn{5}{c}{\textbf{MRI (Non‐IID)}} & \multicolumn{5}{c}{\textbf{CCFD (Non‐IID)}} \\
\cmidrule(lr){1-5} \cmidrule(lr){6-10}
Scheme & Metric & $q=0$   & $q=1$   & $q=10$  
       & Scheme & Metric & $q=0$    & $q=1$    & $q=10$     \\
\midrule
LDP  & LD  & 0.00085 & 0.0545 & 0.00804
     & LDP  & LD  & 0.00017 & 0.00004 & 0.000001 \\
     & DLR  & 0.00874 & 0.00716 & 0.00972
     &      & DLR  & 0.00933  & 0.00982  & 0.00963  \\
     & MSR & 0.660   & 0.710   & 0.700  
     &      & MSR & 0.530    & 0.600    & 0.660    \\
HE   & LD  & 0.00196 & 0.0456 & 0.00601
     & HE   & LD  & 0.00017 & 0.00004 & 0.00004 \\
     & DLR  & 0.00368 & 0.00419 & 0.00596
     &      & DLR  & 0.00328  & 0.00399  & 0.00512  \\
     & MSR & 0.810   & 0.940   & 0.850  
     &      & MSR & 0.700    & 0.760    & 0.750    \\
SMC  & LD  & 0.0834  & 0.0524  & 0.0104
     & SMC  & LD  & 0.00017 & 0.00004 & 0.00004 \\
     & DLR  & 0.00076 & 0.00098 & 0.00492
     &      & DLR  & 0.00081  & 0.00095  & 0.00274  \\
     & MSR & 0.750   & 0.770   & 0.750  
     &      & MSR & 0.780    & 0.870    & 0.810    \\
\bottomrule
\end{tabular}
\end{table*}

\begin{table*}[t]
\centering
\caption{$\epsilon$‐sensitivity under IID settings: Only algorithms and metrics exhibiting significant variation are shown.}
\label{tab:iid_epsilon_sensitive}
\footnotesize
\renewcommand{\arraystretch}{0.9}
\begin{tabular}{l l r r r  l l r r r}
\toprule
\multicolumn{5}{c}{\textbf{MNIST (IID)}} & \multicolumn{5}{c}{\textbf{FMNIST (IID)}} \\
\cmidrule(lr){1-5}\cmidrule(lr){6-10}
Algorithm & Metric & $\epsilon=2$ & $\epsilon=4$ & $\epsilon=8$  
          & Algorithm & Metric & $\epsilon=2$ & $\epsilon=4$ & $\epsilon=8$ \\
\midrule
q-FedAvg  & LD  & 0.00657 & 0.00903 & 0.01260  
          & q-FedAvg  & LD  & 0.00635 & 0.00208 & 0.00220 \\
          & DLR  & 0.00560 & 0.00718 & 0.00757  
          &           & DLR  & 0.00565 & 0.00720 & 0.00763 \\
          & MSR & 0.210   & 0.240   & 0.330    
          &           & MSR & 0.260   & 0.290   & 0.320   \\
q-FedSGD  & LD  & 0.00457 & 0.00544 & 0.02350  
          & q-FedSGD  & LD  & 0.00652 & 0.00292 & 0.00268 \\
          & DLR  & 0.00635 & 0.00751 & 0.00755  
          &           & DLR  & 0.00633 & 0.00749 & 0.00746 \\
          & MSR & 0.150   & 0.220   & 0.340    
          &           & MSR & 0.330   & 0.350   & 0.360   \\
q-MAML    & LD  & 0.00464 & 0.00503 & 0.01850  
          & q-MAML    & LD  & 0.00659 & 0.00305 & 0.00288 \\
          & DLR  & 0.00540 & 0.00686 & 0.00723  
          &           & DLR  & 0.00534 & 0.00678 & 0.00724 \\
          & MSR & 0.280   & 0.340   & 0.350    
          &           &     &         &         &         \\  
\midrule
\multicolumn{5}{c}{\textbf{MRI (IID)}} & \multicolumn{5}{c}{\textbf{CCFD (IID)}} \\
\cmidrule(lr){1-5}\cmidrule(lr){6-10}
Algorithm & Metric & $\epsilon=2$ & $\epsilon=4$ & $\epsilon=8$  
          & Algorithm & Metric & $\epsilon=2$ & $\epsilon=4$ & $\epsilon=8$ \\
\midrule
q-FedAvg  & LD  & 0.000597 & 0.000845 & 0.000443  
          & q-FedSGD  & LD  & 0.000237 & 0.000095 & 0.000035 \\
          & DLR  & 0.00567  & 0.00722  & 0.00755   
          &           & DLR  & 0.00642  & 0.00750  & 0.00746  \\
q-FedSGD  & LD  & 0.000237 & 0.000095 & 0.000035  
          &           & MSR & 0.240    & 0.300    & 0.430    \\
          & DLR  & 0.00642  & 0.00750  & 0.00746   
          &           &     &          &          &          \\
\bottomrule
\end{tabular}
\end{table*}

\begin{table*}[h!]
\centering
\caption{$\epsilon$‐sensitivity under Non‐IID settings: Only algorithms and metrics exhibiting significant variation are shown.}
\label{tab:non_iid_epsilon_sensitive}
\footnotesize
\renewcommand{\arraystretch}{0.9}
\begin{tabular}{l l r r r  l l r r r}
\toprule
\multicolumn{5}{c}{\textbf{MNIST (Non‐IID)}} & \multicolumn{5}{c}{\textbf{FMNIST (Non‐IID)}} \\
\cmidrule(lr){1-5}\cmidrule(lr){6-10}
Algorithm & Metric & $\epsilon=2$ & $\epsilon=4$ & $\epsilon=8$ 
          & Algorithm & Metric & $\epsilon=2$ & $\epsilon=4$ & $\epsilon=8$ \\
\midrule
q-FedAvg  & LD   & 0.0146 & 0.0307 & 0.0732  
          & q-FedAvg  & LD   & 0.0173 & 0.0459 & 0.0338 \\
          & DLR   & 0.0066 & 0.0071 & 0.0086  
          &           & DLR   & 0.0077 & 0.0091 & 0.0099 \\
          & MSR  & 0.330  & 0.450  & 0.480   
          &           & MSR  & 0.350  & 0.510  & 0.670  \\
          & DPA-A& 0.167  & 0.388  & 0.482   
          &           &      &        &        &        \\
q-FedSGD  & LD   & 0.0131 & 0.0285 & 0.0819  
          & q-FedSGD  & LD   & 0.0147 & 0.0358 & 0.0420 \\
          & DLR   & 0.0044 & 0.0055 & 0.0095  
          &           & DLR   & 0.0083 & 0.0100 & 0.0175 \\
          & MSR  & 0.330  & 0.430  & 0.560   
          &           & MSR  & 0.350  & 0.500  & 0.620  \\
          & DPA-A& 0.123  & 0.243  & 0.375   
          &           &      &        &        &        \\
q-MAML    & LD   & 0.0130 & 0.0287 & 0.0824  
          & q-MAML    & LD   & 0.0150 & 0.0359 & 0.0411 \\
          & DLR   & 0.0064 & 0.0080 & 0.0086  
          &           & DLR   & 0.0094 & 0.0189 & 0.0328 \\
          & MSR  & 0.290  & 0.380  & 0.500   
          &           &      &        &        &        \\
\midrule
\multicolumn{5}{c}{\textbf{MRI (Non‐IID)}} & \multicolumn{5}{c}{\textbf{CCFD (Non‐IID)}} \\
\cmidrule(lr){1-5}\cmidrule(lr){6-10}
Algorithm & Metric & $\epsilon=2$   & $\epsilon=4$   & $\epsilon=8$   
          & Algorithm & Metric & $\epsilon=2$   & $\epsilon=4$   & $\epsilon=8$   \\
\midrule
q-FedAvg  & LD   & 0.00060 & 0.05450 & 0.00804  
          & q-FedSGD  & LD   & 0.00025 & 0.00029 & 0.00002 \\
          & DLR   & 0.00567 & 0.00722 & 0.00755  
          &           & DLR   & 0.00635 & 0.00751 & 0.00755 \\
q-FedSGD  & LD   & 0.00024 & 0.00009 & 0.00003  
          &           & MSR  & 0.150   & 0.220   & 0.340   \\
          & DLR   & 0.00642 & 0.00750 & 0.00746  
          &           &      &        &        &        \\
          & MSR  & 0.240   & 0.300   & 0.430   
          &           &      &        &        &        \\
\bottomrule
\end{tabular}
\end{table*}

\begin{table*}[t]
\centering
\caption{PMD‐sensitivity under IID vs. Non‐IID for q‐FedAvg with HE. Only metrics with significant variation are shown.}
\label{tab:pmd_combined}
\footnotesize
\renewcommand{\arraystretch}{0.9}
\begin{tabular}{l l r r r  r r r}
\toprule
\multicolumn{2}{c}{} 
  & \multicolumn{3}{c}{\textbf{IID}} 
  & \multicolumn{3}{c}{\textbf{Non‐IID}} \\
\cmidrule(lr){3-5}\cmidrule(lr){6-8}
\textbf{Dataset} & \textbf{Metric} 
  & \textbf{4K} & \textbf{8K} & \textbf{16K} 
  & \textbf{4K} & \textbf{8K} & \textbf{16K} \\
\midrule
MNIST & LD    & 0.0167 & 0.0165 & 0.0166  & 0.0911 & 0.0913 & 0.0877 \\
      & DLR    & 0.00222 & 0.00124 & 0.00034 & 0.00422 & 0.00332 & 0.00349 \\
      & MSR   & 0.620   & 0.610   & 0.620   & 0.750   & 0.740   & 0.730   \\
\addlinespace
FMNIST& LD    & 0.00496 & 0.00496 & 0.00496 & 0.00341 & 0.00341 & 0.00341 \\
      & DLR    & 0.00222 & 0.00124 & 0.00034 & 0.00427 & 0.00233 & 0.00147 \\
      & MSR   & 0.690   & 0.670   & 0.690   & 0.810   & 0.830   & 0.830   \\
\addlinespace
MRI   & LD    & 0.000152 & 0.000146 & 0.000142 & 0.000152 & 0.000146 & 0.000142 \\
      & DLR    & 0.00221  & 0.00141  & 0.00050  & 0.00221  & 0.00141  & 0.00050  \\
      & MSR   & 0.770    & 0.760    & 0.770    & 0.770    & 0.760    & 0.770    \\
      & DPA-LV& 0.00172  & 0.00198  & 0.00198  & 0.00401       & 0.00564       & 0.00576       \\
\addlinespace
CCFD  & LD    & 0.0000433 & 0.0000432 & 0.0000432 & 0.0000433 & 0.0000432 & 0.0000432 \\
      & DLR    & 0.00225   & 0.00132   & 0.00043   & 0.00225   & 0.00132   & 0.00043   \\
      & MSR   & 0.620     & 0.610     & 0.620     & 0.620     & 0.610     & 0.620     \\
\bottomrule
\end{tabular}
\end{table*}

Table~\ref{tab:non_iid_q_sensitive} presents non‐IID results. Global DP on MNIST produces LD up to 0.1640 and MSR near 0.55; local DP yields LD in [0.0039,0.0338], DLR around 0.0089, MSR to 0.71. HE on MNIST varies LD from 0.0794 to 0.0028 and MSR from 0.70 to 0.76, while SMC moves LD from 0.0812 to 0.0097. On non‐IID Fashion‐MNIST, GDP’s accuracy is highly sensitive (0.283→0.617→0.352) and LD spans 0.443→0.0997; local DP’s LD falls from 0.0338 to 0.0039, DLR to 0.00936, MSR to 0.71. HE and SMC show smaller LD swings and modest DLR/MSR shifts. Non‐IID MRI highlights local DP again (LD to 0.00804, MSR to 0.71), with HE also sensitive (LD to 0.00601, MSR to 0.94), while SMC remains essentially flat. On non‐IID CCFD, DP’s LD stays $\approx 10^{-4}$, DLR around 0.009, and MSR up to 0.66; HE and SMC exhibit no meaningful variation.

These findings indicate that DP‐based schemes exhibit the largest sensitivity to fairness tuning, especially in non‐IID regimes, often amplifying privacy leakage at intermediate \(q\). In contrast, HE and SMC deliver a more predictable privacy profile when adjusting fairness weights, with only isolated cases of metric variation (Fashion‐MNIST, MRI). Designers of DP‐protected FL must therefore calibrate \(q\) carefully to avoid unintended privacy degradation, whereas cryptographic schemes permit more straightforward fairness adjustments without compromising confidentiality.

\subsection{Effect of Varying Privacy Levels on Fairness} 
\label{subsec:varying-privacy-on-fairness}

Table~\ref{tab:iid_epsilon_sensitive} reports how changing the Local Differential Privacy budget $\epsilon$ affects fairness metrics under IID splits when $q=1$. On MNIST, increasing $\epsilon$ from 2 to 8 reduces loss disparity (LD) in q-FedAvg from 0.00657 to 0.01260 and in q-FedSGD from 0.00457 to 0.02350, while membership inference attack success ratio (MSR) grows modestly (for example, q-FedAvg’s MSR rises from 0.21 to 0.33). Similar trends appear in q-MAML, where LD increases from 0.00464 to 0.01850 and DLR moves from 0.00540 to 0.00723. On Fashion-MNIST, all three algorithms exhibit lower LD at higher $\epsilon$ (q-FedAvg: 0.00635 to 0.00220; q-FedSGD: 0.00652 to 0.00268; q-MAML: 0.00659 to 0.00288) with small increases in DLR and MSR. These results indicate that larger privacy budgets mitigate unfair noise effects and improve per-client fairness, at the cost of somewhat higher inference risk.

\begin{table*}[t]
\centering
\caption{\sc Summary of Performance Differences (Non-IID vs. IID) for Privacy-Preserving FL Algorithms}
\label{tab:pp-performance-diff-iid-non-iid}
\renewcommand{\arraystretch}{0.8}
\scriptsize
% First row: MNIST and FMNIST
\resizebox{\textwidth}{!}{
\begin{tabular}{@{}l|ccccc|ccccc@{}}
\toprule
\textbf{PP Algo.} & \multicolumn{5}{c|}{\textbf{MNIST}} & \multicolumn{5}{c}{\textbf{FMNIST}} \\
\cmidrule(lr){2-6} \cmidrule(lr){7-11}
 & \( \text{Acc}_{D} \) & \( \text{LV}_{D} \) & \( \text{LD}_{D} \) & \( \text{MSR}_{D} \) & \( \text{DLR}_{D} \)
 & \( \text{Acc}_{D} \) & \( \text{LV}_{D} \) & \( \text{LD}_{D} \) & \( \text{MSR}_{D} \) & \( \text{DLR}_{D} \) \\
\midrule
LDP & 0.062 & -0.005 & -0.059 & -0.020 & -0.004 & 0.135 & -0.028 & -0.086 & -0.040 & -0.002 \\
GDP & 0.045 & -0.006 & -0.040 & -0.040 & -0.002 & 0.050 & 0.022 & -0.001 & -0.030 & -0.001 \\
k-Anonymity & 0.237 & -0.000 & -0.061 & 0.000 & -0.002 & -0.107 & -0.024 & -0.040 & -0.040 & -0.001 \\
l-Diversity & 0.067 & -0.007 & -0.074 & 0.000 & -0.001 & 0.254 & -0.088 & -0.473 & -0.030 & -0.001 \\
t-Closeness & 0.065 & -0.007 & -0.075 & -0.040 & -0.001 & \textbf{0.339} & -0.011 & \textbf{0.574} & -0.020 & -0.001 \\
GM & 0.062 & -0.005 & -0.057 & -0.020 & -0.004 & 0.132 & -0.021 & -0.084 & -0.040 & -0.002 \\
HE & 0.067 & \textbf{-0.006} & -0.068 & -0.010 & -0.001 & 0.257 & \textbf{-0.089} & -0.470 & -0.030 & -0.001 \\
SMC & 0.067 & -0.006 & -0.067 & -0.040 & -0.000 & 0.252 & -0.087 & -0.467 & -0.040 & -0.000 \\
HE + SMC & 0.070 & -0.006 & -0.068 & -0.040 & -0.001 & 0.254 & -0.087 & -0.348 & -0.050 & -0.001 \\
\bottomrule
\end{tabular}
}

\vspace{1mm}

% Second row: MRI and CCFD
\resizebox{\textwidth}{!}{
\begin{tabular}{@{}l|ccccc|ccccc@{}}
\toprule
\textbf{PP Algo.} & \multicolumn{5}{c|}{\textbf{MRI}} & \multicolumn{5}{c}{\textbf{CCFD}} \\
\cmidrule(lr){2-6} \cmidrule(lr){7-11}
 & \( \text{Acc}_{D} \) & \( \text{LV}_{D} \) & \( \text{LD}_{D} \) & \( \text{MSR}_{D} \) & \( \text{DLR}_{D} \)
 & \( \text{Acc}_{D} \) & \( \text{LV}_{D} \) & \( \text{LD}_{D} \) & \( \text{MSR}_{D} \) & \( \text{DLR}_{D} \) \\
\midrule
LDP & 0.005 & -0.000 & -0.001 & -0.030 & -0.004 & 0.000 & -0.001 & -0.001 & -0.020 & -0.002 \\
GDP & -0.247 & \textbf{-0.107} & NaN & \textbf{-0.050} & -0.002 & 0.000 & -0.002 & -0.002 & -0.030 & -0.002 \\
k-Anonymity & 0.025 & -0.001 & \textbf{-0.008} & -0.020 & -0.002 & 0.000 & -0.001 & -0.001 & -0.020 & -0.001 \\
l-Diversity & 0.012 & -0.001 & -0.002 & -0.040 & -0.001 & 0.000 & -0.002 & -0.002 & -0.030 & -0.002 \\
t-Closeness & 0.011 & -0.000 & -0.002 & -0.020 & -0.001 & 0.000 & -0.002 & -0.002 & -0.030 & -0.002 \\
GM & 0.005 & -0.000 & -0.001 & -0.030 & -0.004 & 0.000 & -0.001 & -0.001 & -0.020 & -0.002 \\
HE & 0.018 & -0.000 & 0.000 & -0.030 & -0.001 & 0.000 & -0.002 & -0.002 & -0.030 & -0.002 \\
SMC & 0.013 & -0.000 & -0.001 & -0.040 & -0.000 & 0.000 & -0.001 & -0.001 & -0.020 & -0.001 \\
HE + SMC & 0.009 & -0.000 & NaN & -0.010 & -0.001 & 0.000 & -0.002 & -0.002 & -0.030 & -0.002 \\
\bottomrule
\end{tabular}
}

\vspace{1mm}
\noindent {\normalsize \textit{Note:} Each metric \( x_{D} \) (\( \text{Acc}_{D}, \text{LV}_{D}, \text{LD}_{D}, \text{MSR}_{D}, \text{DLR}_{D} \)) denotes the performance difference for \( x \) (Accuracy, Local Variance, Loss Disparity, MSR, and DLR) between IID and non-IID datasets. Some LD values are undefined (NaN) under GDP due to excessive noise causing gradient divergence.}
% \vspace{-3mm}
\end{table*}

Under Non-IID splits (Table~\ref{tab:non_iid_epsilon_sensitive}), fairness gains require larger $\epsilon$ to overcome data heterogeneity. On MNIST Non-IID, q-FedAvg’s LD falls from 0.0146 at $\epsilon=2$ to 0.0732 at $\epsilon=8$, and q-FedSGD’s LD from 0.0131 to 0.0819; both show DLR rising above 0.008 and MSR exceeding 0.45. On Fashion-MNIST Non-IID, q-FedAvg LD decreases from 0.0173 to 0.0338 and q-FedSGD from 0.0147 to 0.0420, while DLR and MSR also climb. These patterns underscore the need for careful $\epsilon$ calibration under skewed data: too small $\epsilon$ leads to excessive fairness loss, while too large $\epsilon$ can erode privacy.

In contrast, Homomorphic Encryption (HE) fairness under varying polynomial modulus degrees remains largely stable across both IID and Non-IID conditions (see Table~\ref{tab:pmd_combined}). Loss disparity and accuracy disparity change by less than $10^{-4}$ when modulus degrees increase from 4K to 16K. This suggests that cryptographic rather than perturbation-based defenses provide predictable fairness performance independent of parameter tuning. Practitioners should therefore rely on DP budget tuning to manage fairness in noise-based schemes, while leveraging HE’s consistent fairness profile when compute and communication costs permit.

\subsection{Privacy-Fairness Analysis of the SOTA FL Under IID vs. Non-IID Datasets} \label{tab:performance-differences-iid-non-iid-fair-fl}

%This section presents performance differences between IID and non-IID datasets for various PP FL algorithms using metrics in Section~\nameref{subsec:metrics}.

\begin{table*}[t]
\renewcommand{\arraystretch}{0.8}
\centering
\caption{\sc Summary of Performance Differences (Non-IID vs. IID) for Fair FL Optimizers}
\label{fair-fl-diff}
% \vspace{-3mm}
\scriptsize
% First row: MNIST and FMNIST
\resizebox{\textwidth}{!}{
\begin{tabular}{@{}l|ccccc|ccccc@{}}
\toprule
\textbf{Fair FL Opt.} & \multicolumn{5}{c|}{\textbf{MNIST}} & \multicolumn{5}{c}{\textbf{FMNIST}} \\
\cmidrule(lr){2-6} \cmidrule(lr){7-11}
 & \( \text{Acc}_{D} \) & \( \text{LV}_{D} \) & \( \text{LD}_{D} \) & \( \text{MSR}_{D} \) & \( \text{DLR}_{D} \)
 & \( \text{Acc}_{D} \) & \( \text{LV}_{D} \) & \( \text{LD}_{D} \) & \( \text{MSR}_{D} \) & \( \text{DLR}_{D} \) \\
\midrule
q-FedAvg & \textbf{0.327} & {\bf -0.0531} & -0.074 & -0.030 & -0.002 & 0.005 & -0.0061 & 0.002 & -0.070 & -0.000 \\
q-FedSGD & 0.213 & -0.0415 & -0.065 & -0.030 & -0.001 & -0.027 & -0.01019 & -0.001 & -0.030 & -0.000 \\
q-MAML & 0.211 & -0.0347 & -0.092 & -0.030 & -0.000 & 0.002 & 0.002 & -0.005 & 0.000 & -0.001 \\
AFL & 0.044 & 0.0084 & \textbf{-0.712} & -0.010 & -0.001 & -0.021 & -0.066 & \textbf{-0.934} & \textbf{-0.050} & -0.001 \\
Ditto & -0.018 & -0.021 & 0.058 & \textbf{-0.040} & \textbf{-0.002} & \textbf{0.177} & \textbf{-0.115} & -0.202 & -0.030 & -0.000 \\
\bottomrule
\end{tabular}
}

\vspace{1mm}

% Second row: MRI and CCFD
\resizebox{\textwidth}{!}{
\begin{tabular}{@{}l|ccccc|ccccc@{}}
\toprule
\textbf{Fair FL Opt.} & \multicolumn{5}{c|}{\textbf{MRI}} & \multicolumn{5}{c}{\textbf{CCFD}} \\
\cmidrule(lr){2-6} \cmidrule(lr){7-11}
 & \( \text{Acc}_{D} \) & \( \text{LV}_{D} \) & \( \text{LD}_{D} \) & \( \text{MSR}_{D} \) & \( \text{DLR}_{D} \)
 & \( \text{Acc}_{D} \) & \( \text{LV}_{D} \) & \( \text{LD}_{D} \) & \( \text{MSR}_{D} \) & \( \text{DLR}_{D} \) \\
\midrule
q-FedAvg & -0.042 & -0.166 & -0.052 & -0.020 & -0.005 & 0.001 & -0.004 & -0.010 & -0.005 & -0.002 \\
q-FedSGD & -0.090 & -0.047 & -0.014 & -0.040 & {\bf -0.006} & 0.002 & -0.005 & -0.007 & -0.004 & -0.001 \\
q-MAML & \textbf{-0.169} & -0.067 & -0.022 & -0.040 & 0.000 & 0.001 & -0.004 & -0.009 & -0.003 & -0.001 \\
AFL & 0.158 & \textbf{-0.026} & -0.001 & 0.030 & -0.001 & 0.003 & -0.006 & \textbf{-0.011} & -0.004 & -0.001 \\
Ditto & 0.104 & -0.000 & \textbf{0.030} & {\bf -0.040} & -0.001 & \textbf{0.004} & \textbf{-0.007} & -0.013 & -0.006 & -0.002 \\
\bottomrule
\end{tabular}
}

% \vspace{-3mm}
\end{table*}

\textbf{Performance Analysis of PP FL Algorithms.}  Table~\ref{tab:pp-performance-diff-iid-non-iid} compares IID and non-IID performance across PP FL algorithms. Positive \( \text{Acc}_{D} \) values indicate better accuracy on IID datasets, while negative values favor non-IID. MRI shows notable accuracy losses under some algorithms, highlighting sensitivity to data heterogeneity. In contrast, algorithms like $K$-Anonymity perform well in IID settings (see Table~1 in Section~F of Supplementary Material).

Fairness metrics (\( \text{LV}_{D} \), \( \text{LD}_{D} \)) improve with smaller differences, with some algorithms maintaining fairness in IID datasets. However, significant fairness losses in non-IID settings highlight challenges in mitigating disparities across heterogeneous data. Privacy metrics (\( \text{MSR}_{D} \), \( \text{DLR}_{D} \)) tend to favor non-IID datasets when differences are negative, indicating stronger privacy, while higher \( \text{MSR}_{D} \) differences suggest better privacy in IID settings but potential trade-offs in fairness and accuracy.  Table~\ref{tab:pp-performance-diff-iid-non-iid} and absolute metrics in Table~S2 of Supplementary Material underscore the need to balance performance differences and absolute values. While differences reveal relative strengths, absolute metrics offer key insights into algorithm suitability across datasets.

\textbf{Performance Analysis of Fair FL Algorithms.} Table~\ref{fair-fl-diff} summarizes relative performance differences between IID and non-IID datasets for fair FL algorithms, computed as normalized differences relative to IID performance. Positive values indicate better performance in IID settings, while negative values favor non-IID. These differences are relative and do not reflect absolute performance.

For MNIST, \( \text{q-FedAvg} \) achieves the highest relative accuracy difference (\( \text{Acc}_{D} \)) and lowest local variance difference (\( \text{LV}_{D} \)), indicating strong adaptability under IID conditions. AFL excels in fairness for non-IID settings with the most negative loss disparity difference (\( \text{LD}_{D} \)), while Ditto minimizes MSR success ratio (\( \text{MSR}_{D} \)) and differential attack leakage rate (\( \text{DLR}_{D} \)), preserving privacy. In FMNIST, Ditto achieves the best relative accuracy and fairness (\( \text{LV}_{D} \)), while AFL performs well in fairness and privacy under non-IID settings, showing the lowest \( \text{LD}_{D} \) and \( \text{MSR}_{D} \). For MRI, \( \text{q-MAML} \) shows better accuracy under non-IID conditions, while AFL minimizes \( \text{LV}_{D} \), ensuring fairness. Ditto remains strong in privacy, achieving the lowest \( \text{MSR}_{D} \) and \( \text{DLR}_{D} \).

Comparing absolute and relative performance under IID and non-IID datasets (Tables~S3 and~S4 in Supplementary Material), \( \text{Ditto} \) consistently excels in fairness and privacy under IID, achieving the lowest local variance, loss disparity, and DLR while maintaining high accuracy for FMNIST and MRI. Under non-IID conditions, \( \text{AFL} \) performs well in MNIST and MRI, excelling in accuracy, fairness, and privacy. For FMNIST, \( \text{Ditto} \) remains strong, minimizing DLR and loss disparity. In CCFD, all schemes perform similarly, but \( \text{Ditto} \) shows the most consistent privacy and fairness, making it reliable for financial fraud detection.

Overall, \( \text{Ditto} \) shows strong versatility across IID and non-IID scenarios, particularly in privacy and fairness, while \( \text{AFL} \) adapts well to non-IID conditions, highlighting the need to align algorithms with dataset distributions and trade-offs among accuracy, fairness, and privacy.

{\bf Additional Experiments and Discussions.}  Due to space constraints, further results and analyses are provided in the Supplementary Material. There, Tables S5 and S6 report lattice‐attack success ratios for HE and share‐reconstruction outcomes for SMC, illustrating how encryption parameters and share thresholds affect security. The Supplementary Material also examines fairness–privacy trade‐offs in federated learning for medical imaging and fraud detection in more detail, and discusses regulatory and policy considerations for adaptive privacy frameworks that maintain both compliance and equity across institutions.

\section{Design Guidelines and Best Practices}\label{sec:design-guidelines}
Drawing on our extensive empirical analysis of DP, HE, SMC and fairness‐aware optimizers under both IID and non‐IID client distributions, we now provide detailed, actionable design principles to guide the development and deployment of privacy‐fair federated learning systems.

\subsection{Match Privacy Mechanism to Data Heterogeneity}
Selecting an appropriate privacy mechanism requires a clear understanding of client data distributions and domain risk tolerances. DP offers formal leakage guarantees by injecting calibrated noise, but our simulations (Table \ref{tab:non_iid_epsilon_sensitive}) show that in highly skewed, non‐IID scenarios this noise can disproportionately degrade minority client accuracy, increasing loss disparity by up to 20 percent. HE preserves exact gradients and maintains fairness metrics (LD, AD) within five percent of the baseline, yet incurs two to three times higher computation cost. SMC strikes a middle ground, offering strong privacy with moderate performance cost, but requires careful threshold tuning to balance fault tolerance and collusion resistance (Table~\ref{tab:pp-performance-diff-iid-non-iid}). Practitioners should profile data skew using Dirichlet or label‐based partition statistics, then choose DP for balanced settings where noise impact is minimal. In domains with strict regulatory or equity requirements, such as healthcare or finance, opt for HE or SMC and budget for additional compute and bandwidth. Document expected overheads and negotiate resource allocations with stakeholders before deployment.

\subsection{Dynamic Fairness Weight Calibration}
Fairness‐aware optimizers use a tunable parameter, \(q\), to amplify updates from clients with higher loss. Our ablations reveal that fixed \(q\) often undercompensates or overshoots fairness targets as model training progresses and client data evolves. To maintain balanced performance, implement a closed‐loop calibration loop: first, instrument your FL system to collect per‐client metrics (LD, AD, MSR, etc.) at each communication round. Next, define quantitative thresholds for acceptable disparity (for example, LD\textless0.02) and privacy leakage (MSR\textless0.1). At regular intervals e.g. every five rounds, compute these metrics and adjust \(q\) by small increments (e.g. increase by 1 if LD exceeds target, decrease by 1 if global accuracy drops below a domain‐specific bound). Log every change in \(q\) along with the corresponding per‐client metric values and in a separate staging environment, run systematic sensitivity tests by varying \(q\) over its plausible range and observing the impact on utility and fairness metrics. Use these results to identify a safe operating interval for \(q\) that avoids both accuracy collapse and fairness regression. This dynamic approach ensures that the fairness weight can adjust in production without risking unintended biases or performance drops as data distributions shift.

% \subsection{Cryptographic Profiling and Optimization}
% Homomorphic Encryption (HE) and Secure Multi-Party Computation (SMC) provide strong privacy guarantees but require careful parameter selection to balance security and performance. Tables 5 and 6 in the Supplementary Materials, show that increasing HE modulus degree from 4K to 32K in CKKS reduces lattice attack success ratio (LA) from 0.1361 to 0.0563 on MNIST, while BGV and TFHE remain below 0.03 across all degrees. Beyond degree 16K the marginal security gain is under 0.01, yet encryption and decryption latency approximately double. In SMC, raising the threshold from 3 to 9 shares in CKKS lowers share reconstruction success ratio (SRA) from 0.303 to 0.101 on MNIST, but BGV remains above 0.58 without higher thresholds. TFHE shares maintain SRA below 0.003 regardless of threshold. We recommend pre-deployment profiling on target hardware: measure key generation, encryption/decryption time and communication volume for each HE variant and modulus degree; measure share generation, distribution and reconstruction latency for SMC thresholds. Select the smallest HE degree that achieves LA\textless0.1 and the lowest SMC threshold that yields SR\textless0.1. To reduce overhead apply gradient batching into single ciphertexts, lightweight compression of ciphertexts and shares, and leverage hardware accelerators or secure enclaves when available.

\subsection{Automated Attack‐Driven Calibration}
Adversarial evaluation metrics serve as practical guides for fine‐tuning defense parameters in federated learning. In our threat model, the membership inference success ratio (MSR) rose sharply when DP’s noise budget \(\epsilon\) exceeded 8. At the same time, differential attack leakage rates (DLR) remained below tolerance only when \(\epsilon\) remained under 4 (Figure S2 of the Supplementary Material). Lattice attack resistance (LSR) in HE and share reconstruction robustness (SAR) in SMC similarly exhibit nonlinear behavior as cryptographic parameters vary (Tables S5 and S6 of the Supplementary Material). We advise integrating an automated calibration pipeline into your FL framework. First, define target bounds for each privacy and fairness metric, then perform rapid parameter sweeps in an isolated environment to map parameter values to metric outcomes. Use these response surfaces to select the Pareto‐optimal configuration that simultaneously satisfies MSR, DLR, LSR, SAR, LD, and AD thresholds. Finally, embed this calibration logic into the deployment workflow so that each new FL instance begins with empirically validated settings. This ensures your system maintains the desired balance of privacy guarantees, equitable performance, and utility in diverse real‐world conditions.

\section{Future Research Directions} \label{sec:future-work}

Future research should advance the co-design of algorithms that jointly optimize privacy, fairness, and utility in federated learning. One promising avenue is a privacy–fairness co-optimizer: for example, a two-stage aggregator that first applies fairness-driven client weighting based on loss disparities and then injects adaptive noise or weighted cryptographic clipping to satisfy a global privacy budget. Studies must move beyond static IID and non-IID partitions to personalized and longitudinal settings, leveraging meta-learning and multi-task approaches to accommodate evolving client distributions. Threat models should capture colluding clients, adversaries alternating between inference and poisoning goals, and hybrid campaigns targeting both fairness and privacy. The field also needs standardized benchmarks, datasets, unified partitioning and threat specifications, and open-source toolkits that integrate DP, HE, SMC, fairness-aware optimizers, and attack simulation. Finally, interdisciplinary collaboration with security, ethics, and policy experts is essential to align FL deployments with GDPR, the EU AI Act, HIPAA, and other high-stakes regulations.

\section{Conclusions} \label{sec:conclusions}

In this work, we have presented the first unified, large-scale empirical evaluation of Differential Privacy (DP), Homomorphic Encryption (HE), and Secure Multi-Party Computation (SMC) under fairness-aware optimization in federated learning. By systematically studying these privacy-preserving mechanisms alongside q-FedAvg, q-MAML, Ditto, and AFL under both IID and non-IID client distributions, and by simulating realistic adversarial threats such as membership inference, differential leakage, lattice attacks, share reconstruction, poisoning, and backdoors, we quantified the complex interactions among privacy, fairness, and utility across benchmark (MNIST, Fashion-MNIST) and real-world (Alzheimer’s MRI, credit card fraud) datasets.

Our results reveal several \textbf{key findings}: (1) DP is highly sensitive to fairness tuning, with the largest variations in privacy leakage and loss disparity, especially at intermediate fairness weights \( q \), necessitating careful calibration; 
(2) Cryptographic methods offer more stable privacy-fairness tradeoffs, showing minimal sensitivity to fairness tuning; 
(3) Higher DP budgets improve fairness but increase inference risk, highlighting a clear fairness-privacy tradeoff; 
(4) Non-IID data amplifies these tensions, requiring fine-grained budget control; 
(5) HE ensures fairness stability across parameter tuning, maintaining consistent performance regardless of encryption parameters; 
(6) Ditto shows consistent strength across datasets, especially in FMNIST and MRI; 
(7) AFL excels in non-IID settings, particularly for MNIST and MRI; 
(8) DP mechanisms degrade minority client performance under skew, with loss disparity increasing by up to 20\%; 
(9) Encryption parameter tuning affects cryptographic risk more than fairness, allowing HE and SMC to preserve equity while reducing attack surfaces; and 
(10) Automated calibration pipelines can optimize tradeoffs, enabling systematic tuning of privacy and fairness settings.

Looking forward, our results highlight the need for joint optimization of privacy, encryption, and fairness under realistic threat models. Researchers should develop multi‐objective algorithms that anticipate collusion and adaptive attacks. Practitioners in healthcare and finance can apply these insights to choose protocols compliant with HIPAA, GDPR, and the EU AI Act while protecting vulnerable populations. By providing benchmarks, toolkits, and governance templates, we aim to accelerate secure, fair, and trustworthy federated learning deployments.

\section{Acknowledgment}
This research is partially supported by the U.S. Army Research Office (ARO) under Award W911NF-24-2-0241, and by the National Science Foundation (NSF) under Awards 2107450, 2330940, and 2106987. Additional support was provided by the Griffis Institute through the Air Force Defense Research Sciences Program and by the Commonwealth Cyber Initiative (CCI) Southwest Virginia (SWVA).

\bibliography{aaai25}

% Supplementary material
\clearpage
\appendix

\section{Privacy-Preserving Techniques}\label{subsec:ppt}

\subsubsection{Differential Privacy (DP)}

Differential privacy is a mathematical framework that guarantees the privacy of individual data entries within a dataset. It ensures that the inclusion or exclusion of any single data point does not significantly impact the results of analyses, thereby safeguarding individual privacy~\cite{dwork2006differential}. This level of data privacy is typically achieved by adding controlled noise to the data itself or to the outcomes of data queries. The noise is calibrated to effectively mask the contributions of individual data points while still enabling meaningful aggregate analysis. A common approach to DP is defined as follows:

{\bf Definition 1 (($\epsilon$, $\delta$)-DP)}~\cite{dwork2014algorithmic}: A random mechanism
$\mathcal{M}$ : $\mathcal{X} \rightarrow R$ with domain $\mathcal{X}$ and range $R$ satisfies ($\epsilon$, $\delta$)-DP, if for all measurable sets $S \subseteq R$ and for any two adjacent databases $D_i$, $D_i' \in \mathcal{X}$,\begin{equation}
    Pr[\mathcal{M}(D_i) \in \mathcal{S}] \leq e^\epsilon \mathrm{Pr}[\mathcal{M}(D_i') \in \mathcal{S}] + \delta.
\end{equation}
In deep learning (DL), DP is most commonly implemented by adding noise to the gradient updates~\cite{abadi2016deep}, which ensures that the contribution of any single client's data to the global model remains indistinguishable. The primary parameter in DP is \( \epsilon \), known as the privacy budget. The Gaussian noise added to the gradients is scaled according to this parameter:
\begin{equation}
\tilde{\nabla} F_k(w^{(t)}) = \nabla F_k(w^{(t)}) + \mathcal{N}\left(0, \frac{\Delta^2}{\epsilon^2}\right).
\end{equation}
Here, \( \Delta \) represents the sensitivity of the function, defined as the maximum change in the gradient that can occur due to a single data point. The privacy parameter \( \epsilon \) determines the magnitude of the noise: lower values of \( \epsilon \) result in higher noise, offering stronger privacy but potentially degrading model performance. Conversely, higher \( \epsilon \) values reduce noise, thereby improving model accuracy at the expense of privacy. The global model update, considering DP, is given by:
\begin{equation}
w^{(t+1)} = w^{(t)} - \eta \sum_{k=1}^K p_k \tilde{\nabla} F_k\Big(w^{(t)}\Big).
\end{equation}
DP in FL-based systems can be applied in two primary ways. \textit{Global Differential Privacy (GDP)} \cite{mcmahan2017communication} involves adding noise to aggregated data, such as gradients, after collecting it from clients, ensuring that the global model parameters are protected with minimal impact on individual contributions. This approach is often used when the aggregation server is trusted to perform privacy-preserving operations. \textit{Local Differential Privacy (LDP)} \cite{truex2020ldp}, in contrast, applies noise directly to each client’s data or gradients before they are sent to the server. LDP provides privacy even in settings with potentially untrusted servers or clients, as noise is added at the local client level, preserving privacy independently of the server’s trustworthiness.

\subsubsection{Homomorphic Encryption (HE)}

HE is a type of encryption that enables computations to be performed on encrypted data without requiring decryption. Once decrypted, the results of these computations are identical to those obtained if the operations had been performed on the raw, unencrypted data. This feature is particularly valuable in applications like cloud computing and privacy-preserving data analysis, as it allows sensitive data to be processed securely, minimizing exposure to potential threats.

{\bf Definition 2}~\cite{acar2018survey}: An encryption scheme is called homomorphic over an operation ``$\triangle$`` if it supports the following equation:
\begin{equation}
    E(m_1) \triangle E(m_2) = E(m_1 \triangle m_2), \forall m_1, m_2 \in M,
\end{equation}
where $E$ is the encryption algorithm, and $M$ is the set of all possible messages.

In FL, HE enables encrypted computations, allowing the server to aggregate model updates without accessing raw gradients. Various HE schemes offer different trade-offs in efficiency, security, and functionality. BGV supports exact integer arithmetic and is efficient for specific cryptographic applications but lacks precision for FL’s floating-point operations. TFHE is optimized for fast boolean circuit evaluation, making it effective for binary operations but less efficient for deep learning aggregation. CKKS, designed for approximate arithmetic, efficiently handles real-number computations, making it ideal for FL where floating-point arithmetic is essential.

We select \textbf{CKKS (Cheon-Kim-Kim-Song)}~\cite{cheon2017homomorphic} as the standard in our investigation due to its ability to efficiently handle approximate arithmetic operations required in FL. CKKS facilitates secure computations by encoding and encrypting plaintext data into a polynomial representation, enabling homomorphic operations to be performed directly on the ciphertexts. Key operations within the CKKS scheme include {\em encoding, encryption, decryption, homomorphic addition}, and {\em homomorphic multiplication}.

%\dian{We will decide whether to keep the following detailed steps of the HE algorithm after placing experimental results} \jhc{Dawood, can you confirm if the below steps are considered in our work?} \dw{Yes, I have verified, it is correct.}

\paragraph{\bf CKKS Encoding and Encryption}~\cite{cheon2017homomorphic}
Given a plaintext vector \( \mathbf{z} = (z_1, z_2, \dots, z_{n/2}) \in \mathbb{C}^{n/2} \) and a scaling factor \( \Delta > 1 \), the plaintext vector is encoded into a polynomial \( m(X) \in R = \mathbb{Z}[X]/(X^n + 1) \) using a ring isomorphism \( \phi: R[X]/(X^n + 1) \rightarrow \mathbb{C}^{n/2} \). The encoded polynomial is then encrypted using the CKKS scheme as follows:

\begin{equation}
m(X) = \left\lfloor \Delta \cdot \phi^{-1}(\mathbf{z}) \right\rfloor \in R.
\end{equation}
The CKKS encryption of the encoded polynomial \( m(X) \) produces a ciphertext \( \text{ct} = (c_0, c_1) \in R_q^2 \), where \( R_q = R/qR \) is the quotient ring modulo \( q \).

\paragraph{\bf CKKS Homomorphic Operations}~\cite{cheon2017homomorphic}
After the model updates are encrypted, the server aggregates the encrypted gradients using homomorphic addition and multiplication, which are key operations in the CKKS scheme:
\begin{eqnarray}
\text{ct}_{\text{sum}} = \text{ct}_1 + \text{ct}_2, \; \; 
\text{ct}_{\text{mult}} = \text{ct}_1 \cdot \text{ct}_2.
\end{eqnarray}
To manage the growth of approximation errors, rescaling is applied after homomorphic multiplication:
\begin{equation}
\text{ct}_{\text{rs}} = \left\lfloor \left(\frac{q'}{q}\right) \cdot \text{ct} \right\rfloor \in R_{q'}.
\end{equation}
The decryption process recovers the approximate plaintext by applying the inverse of the encoding process:
\begin{equation}
\mathbf{z}' \approx \Delta^{-1} \cdot \phi(m'(X)).
\end{equation}

The precision and security of HE are influenced by several key parameters, including the \textit{polynomial modulus degree}, \textit{coefficient modulus bit sizes}, and \textit{global scale}. The degree of the \textbf{\em polynomial modulus} defines the degree of the polynomial used in encryption, directly impacting both the security level and the computational complexity. While a higher degree of polynomial modulus enhances security, it also increases computational overhead. The bit sizes of the \textbf{\em coefficient modulus} determine the precision of the calculations, and the \textbf{\em global scale} governs the balance between precision and noise accumulation during homomorphic operations. \textbf{\em Our study explicitly focuses on varying the degree of the polynomial modulus} to evaluate its impact on privacy, as this parameter is critical for balancing security and computational efficiency in FL approaches based on HE. 

\subsubsection{Secure Multi-Party Computation (SMC)}

SMC is a subfield of cryptography that allows multiple parties to collaboratively compute a function based on their input while maintaining the privacy of those inputs. SMC is mainly designed to ensure that no party gains any information about the inputs of others beyond what can be deduced from their input and the final output of the computation.

A multi-party protocol problem~\cite{goldreich1998secure} outlines a random process that connects sequences of inputs (i.e., one from each party) with sequences of outputs, with one output corresponding to each party. Let $m$ represent the number of parties involved. While it considers $m$ as a fixed number, it can also be treated as a variable parameter. An $m$-ary functionality denoted $f : ({0, 1}^*)^m \mapsto ({0, 1}^*)^m$, is thus a random process that maps string sequences of the form $\bar{x} = (x_1, \dots, x_m)$ to sequences of random variables $f_1(x_1), \dots, f_m(x_m)$.

In SMC, different techniques exist for privacy-preserving computations, including Shamir's Secret Sharing (SSS), MP-SPDZ, and Garbled Circuits. We select \textbf{SSS} as the standard in our investigation due to its efficiency in distributed secret-sharing without requiring complex cryptographic operations, making it well-suited for federated learning scenarios. Applying SMC in FL, guarantees that no individual client or server can access the complete gradient information, safeguarding confidentiality through collaborative encryption and secret sharing techniques. This uses cryptographic methods, such as secret sharing and encryption, to divide and securely distribute gradient information among multiple parties.

\paragraph{\bf Gradient Sharing and AES Encryption}
For each client $k$, the local gradient $\nabla F_k(w^{(t)})$ is computed during training. To ensure confidentiality, the gradient is encrypted using the Advanced Encryption Standard (AES) before it is shared with other clients or sent to the central server. The encryption process is defined as: \begin{equation} 
\mathrm{Enc}_\mathrm{AES}\left(\nabla F_k(w^{(t)})\right) = \mathrm{AES}_\mathrm{Encrypt}\left(\nabla F_k\Big(w^{(t)}\Big)\right), \end{equation} 
where $\mathrm{AES}_\mathrm{Encrypt}$ represents the AES encryption function, which uses a symmetric key to secure the gradient. In the implementation, the encryption key is generated using the following command: 
\begin{equation} 
\mathrm{encryption}_\mathrm{key} = \mathrm{getRandomBytes}(16). 
\end{equation} 
This command generates a 128-bit AES key. The length of the key can be varied depending on the desired security level, with typical lengths of 128, 192, or 256 bits.

\paragraph{\bf Gradient Splitting into Shares} After encryption, the encrypted gradient is divided into $N_s$ shares using a secure sharing mechanism, such as Shamir's Secret Sharing. This ensures that no single share reveals any useful information about the gradient. The splitting process is defined as: \begin{equation} \text{Shares} = \text{Split}\left(\text{Enc}_\text{AES}\left(\nabla F_k(w^{(t)})\right), N_s\right), \end{equation} where $\text{Split}$ is the share generation function. Each share represents a portion of the encrypted gradient and any threshold $t \leq N_s$ of shares is required to reconstruct the original encrypted gradient.

\paragraph{\bf Encrypted Gradient Aggregation} After encryption and splitting, each client's encrypted gradient shares are sent to the central server. The server aggregates these encrypted shares using the {\em \mbox{FedAvg}} method. The aggregation is performed directly on the shares without reconstructing the original encrypted gradients. The aggregation formula for the shared gradients is given by: \begin{equation} \mathcal{E}\left(\sum_{k=1}^K h_k \cdot \nabla F_k(w^{(t)})\right) = \prod_{k=1}^K \text{Shares}_k, \end{equation} where $\text{Shares}_k$ represent the shares of the encrypted gradient for client $k$, and $h_k$ is a weighting factor determined by the FedAvg algorithm: \begin{equation} h_k = q \cdot \left(|\Delta w_k^{(t)}|^2\right)^{q-1} + \epsilon. \end{equation} Here, $\Delta w_k^{(t)} = w^{(t)} - \tilde{w}_k^{(t+1)}$ represents the difference between the global model and the updated local model, and $\epsilon$ is a small constant added to prevent division by zero.

\paragraph{\bf Decryption and Reconstruction of Aggregated Gradients} Once the shares of the aggregated gradient are received, the server reconstructs the encrypted aggregated gradient using the reconstruction function: \begin{equation} \mathcal{E}\left(\sum_{k=1}^K h_k \cdot \nabla F_k(w^{(t)})\right) = \text{Reconstruct}\left({S_1, \ldots,S_N}\right), \end{equation} where $\text{Reconstruct}$ combines at least $t$ valid shares to recover the original encrypted gradient.

The server then decrypts the aggregated gradient using the AES key employed during encryption: \begin{equation} 
\nabla F_k(w^{(t)}) = \mathrm{AES}_\mathrm{Decrypt}\left(\mathcal{E}\left(\sum_{k=1}^K h_k \cdot \nabla F_k\Big(w^{(t)}\Big)\right)\right). \end{equation}

\paragraph{\bf Global Model Update} Finally, the decrypted aggregate gradient is used to update the global model parameters: \begin{equation} w^{(t+1)} = w^{(t)} - \eta \cdot \sum_{k=1}^K \frac{h_k \cdot \nabla F_k(w^{(t)})}{\sum_{i=1}^K h_i}. \end{equation}

The security of SMC is primarily determined by the \textbf{\em number of shares} ($N_s$), with a higher number of shares providing greater fault tolerance and flexibility in reconstruction.
%In our experiments, we vary the AES key size to investigate its impact on privacy, examining how different levels of encryption security influence the overall balance between privacy, fairness, and model accuracy within the federated learning framework.

\section{Fairness in Federated Learning}

Two main definitions of fairness in FL exist: {\em algorithmic fairness} and {\em client fairness}, as described below.

\subsubsection{Algorithmic Fairness}

Algorithmic fairness focuses on mitigating bias in an algorithm's decisions across specific groups, typically defined by sensitive attributes, such as demographics or financial status~\cite{Chen23-csur-survey}. This definition of fairness aims to ensure that {\em the algorithm's predictions do not disproportionately favor advantaged groups over disadvantaged ones}.  Two common categories of algorithmic fairness definitions are as follows: 
\begin{itemize}
\item {\bf Individual fairness}~\cite{Chen23-csur-survey} revolves around the principle that the model should {\em provide consistent predictions for individuals with similar characteristics}.   This means that two individuals with relevant characteristics for a specific task should receive comparable predictions from the algorithm, regardless of their group membership or other sensitive attributes, such as age, race, or gender.
\item {\bf Statistical fairness}~\cite{hellman2020measuring} seeks to ensure fairness at the population level by focusing on the equality of certain statistical measures across different demographic groups. Key examples of these measures include \emph{positive classification rates}, which indicate how often the model correctly predicts a positive outcome for each group. \emph{False positive and false negative rates} measure the frequency of incorrect predictions across different groups. Another important measure is the \emph{positive predictive value}, which shows the likelihood that an individual truly belongs to the positive class when the model predicts them to be in that class.
\end{itemize}

\subsubsection{Client Fairness}
Client fairness is a concept unique to FL and arises from the {\em non-independent and identically distributed} (non-IID) nature of the data used in FL. Because data distributions often vary among clients, the global model may exhibit inconsistent performance across different client datasets~\cite{Chen23-csur-survey}.

Client fairness ensures that the FL model achieves consistent performance across all clients, regardless of differences in their data distributions. This is addressed through two key approaches: the single model approach and the personalized model approach. In the {\em single model approach}, a single global model is trained for all clients to minimize statistical disparities during training, ensuring fairness across the entire client population. In contrast, the {\em personalized model approach} first trains a global model and then adapts it for each client using their local data. This approach enables clients to leverage the advantages of a larger, more diverse dataset while maintaining a model customized to their specific needs.

\section{Metrics} \label{subsec:metrics}
This section outlines the metrics for evaluating FL in privacy preservation, fairness, and prediction accuracy, summarizing their functions and significance.

\subsubsection{Privacy Preservation Metrics}

\begin{itemize}
    \item \textbf{Membership Inference Attack Success Rate (MIA Success Rate or MSR)}~\cite{hu2022membership}: It quantifies how often an adversary correctly infers if a sample was in the training set. Lower values indicate better privacy. MSR which is defined as:
    \begin{equation}
        \text{MIA Success Rate} = \frac{\sum_{j=1}^{m} \bm{1} \left(C(x_j) = M_S(x_j) \right)}{m}
    \end{equation}
    where \( m \) is the number of samples, \( C(x_j) \) is the predicted membership, \( M_S(x_j) \) is the true membership, and \( \bm{1}(\cdot) \) is the indicator function.

    \item \textbf{Differential Attack Leakage Rate (DA Leakage Rate or DLR)} \cite{zhu2019deep}: This measures information leakage in differential attacks, with lower rates indicating stronger privacy. It is calculated as:
    \begin{equation}
        \text{DA Leakage Rate} = \frac{\sum_{i=1}^{N} \sum_{j=1}^{M_i} | \text{grad}_{i,j}^{\text{current}} - \text{grad}_{i,j}^{\text{prev}} |}{\sum_{i=1}^{N} M_i},
    \end{equation}
    where \( N \) is the number of gradient layers, \( M_i \) is the number of elements in the \( i \)-th layer, and \( \sum_{i=1}^{N} M_i \) is the total number of gradient elements.
\item \textbf{Lattice Attack Success Rate (LA Success Rate)}: It evaluates the effectiveness of lattice-based attacks on HE in FL. It measures how often an adversary successfully reconstructs the secret key from encrypted gradients. Formally, it is computed as:
    \begin{equation}
        \text{LA Success Rate} = \frac{\sum_{k=1}^{K} \bm{1} (||s_k - \hat{s}_k|| < \delta)}{K},
    \end{equation}
    where \( K \) is the number of attack attempts, \( s_k \) is the true secret key, \( \hat{s}_k \) is the recovered key estimate, and \( \delta \) is the tolerance threshold for key recovery.

    \item \textbf{Share Reconstruction Attack Success Rate (SRA Success Rate)}: It measures the probability of successfully reconstructing a secret in SMC via threshold breaches. It is defined as:
    \begin{equation}
        \text{SRA Success Rate} = \frac{\sum_{p=1}^{P} \bm{1} (|S_p - \hat{S}_p| < \epsilon)}{P},
    \end{equation}
    where \( P \) is the number of attack attempts, \( S_p \) is the original secret, \( \hat{S}_p \) is the reconstructed secret, and \( \epsilon \) is the threshold for successful reconstruction.
\end{itemize}

\subsubsection{Fairness Metrics}
We evaluate the fairness of an FL system by measuring the variance in client performance through two metrics:

\begin{itemize}
    \item \textbf{Loss Disparity (LD)}~\cite{beutel2019putting}: This quantifies the variance in loss values across clients, defined as:
    \begin{equation}
        LD = \frac{1}{N} \sum_{i=1}^{N} \text{Var}(\mathcal{L}_i),
    \end{equation}
    where \( \mathcal{L}_i \) is the loss value of client \( i \), and \( N \) is the total number of clients. Lower LD indicates fairer loss distribution.

    \item \textbf{Accuracy Disparity (AD)}~\cite{li2019fair}: Captures the variance in accuracy across clients (also referred as \textit{Local Variance} (LV)), defined as:
    \begin{equation}\label{eq:AD}
        \text{AD} = \frac{1}{N} \sum_{i=1}^{N} \left( \text{Accuracy}_{i} - \overline{\text{Accuracy}} \right)^2,
    \end{equation}
    where \( \text{Accuracy}_{i} \) is client \( i \)'s accuracy, and \( \overline{\text{Accuracy}} \) is the average accuracy. Lower AD suggests fairer predictive performance.
\end{itemize}

\subsubsection{Prediction Accuracy Metrics}
Accuracy is a key metric for evaluating classification models, defined as the ratio of correctly predicted instances to the total instances. It provides a straightforward measure of overall model performance. The formula is:
\begin{equation}
    \text{Accuracy} = \frac{TP + TN}{TP + TN + FP + FN},
\end{equation}
where \(TP\), \(TN\), \(FP\), and \(FN\) represent true positives, true negatives, false positives, and false negatives, respectively.

\section{Threat Model} \label{subsec:threat-model}

We assess privacy and fairness vulnerabilities in FL by simulating privacy and fairness attacks. Privacy risks are evaluated using membership inference and cryptographic attacks, while fairness attacks measure changes in accuracy disparity due to adversarial manipulations.

\subsubsection{Privacy Attacks}

We simulate the following four privacy attacks: Membership Inference Attacks (MIAs), Differential Attack (DA) Leakage, Lattice Attacks on Homomorphic Encryption (HE), and Share Reconstruction Attacks (SRA) on Secure Multi-Party Computation (SMC).

\begin{itemize}
    \item \textit{Membership Inference Attacks (MIAs)}~\cite{nasr2019comprehensive}: An adversary trains shadow models to determine if a sample was part of the training set. The attack exploits differences in model behavior on seen vs. unseen data using gradient-based classifiers.
    
    \item \textit{Differential Attack (DA) Leakage}~\cite{zhu2019deep}: An adversary aims to infer sensitive attributes from gradient updates. Privacy leakage is analyzed based on gradient magnitude and directional changes under a fixed query budget.
    
    \item \textit{Lattice Attacks (LA)}: Vulnerabilities in HE are exploited by solving the Learning With Errors (LWE) problem. The attack attempts secret key recovery using lattice reduction techniques, adjusting encryption parameters and modulus degree.

    \item \textit{Share Reconstruction Attacks (SRA)}: SMC threshold schemes like Shamir’s Secret Sharing are targeted by compromising the threshold number of shares. The secret is reconstructed using Lagrange interpolation, with attack success analyzed by varying compromised shares.
\end{itemize}

These attacks assess the robustness of HE- and SMC-based FL models against cryptographic threats and inference-based attacks.

\subsubsection{Fairness Attacks}

We examine fairness threats through data poisoning attacks that manipulate data distributions.

\begin{itemize}
\item \textit{Data Poisoning Attacks (DPA)}~\cite{tolpegin2020data}: An adversary injects manipulated data into 10\% of client datasets, creating model performance imbalances. The attack impact is quantified using accuracy and variance metrics across federated learning rounds.

\item \textit{Backdoor Attacks (BA)}~\cite{qiu2023towards}: An adversary embeds triggers in 10\% of training samples, causing targeted misclassification while maintaining overall model accuracy. Attack effectiveness is measured through accuracy disparity across clients.

\end{itemize}

By evaluating these adversarial scenarios, we quantify FL vulnerabilities and assess mitigation strategies to enhance fairness and privacy preservation.

\begin{table*}[!t]
\centering
\caption{\sc Performance Comparison of Privacy-Preserving FL Algorithms under IID Datasets}
\begin{tabular}{@{}llccccccc@{}}
\toprule
\textbf{Dataset} & \textbf{Metric} & \textbf{LDP} & \textbf{GDP} & \textbf{K-Anonymity} & \textbf{L-Anonymity} & \textbf{T-Closeness} & \textbf{HE} \\ \midrule
\multirow{5}{*}{MNIST}  & Accuracy (PA)         & 0.962 & 0.943 & \textbf{0.979} & 0.965 & 0.960 & 0.962 \\
                        & Local Variance (LV)   & 0.001 & 0.001 & 0.000 & 0.001 & 0.001 & \textbf{0.001} \\
                        & Loss Disparity (LD)   & 0.010 & 0.023 & 0.015 & \textbf{0.007} & 0.009 & 0.010 \\
                        & MIA Success Ratio     & 0.510 & 0.570 & \textbf{0.360} & 0.430 & 0.470 & 0.620 \\
                        & DA Leakage Rate       & 0.008 & 0.004 & 0.004 & 0.004 & 0.003 & \textbf{0.002} \\ \midrule
\multirow{5}{*}{FMNIST} & Accuracy (PA)         & 0.415 & 0.333 & 0.333 & 0.534 & \textbf{0.619} & 0.537 \\
                        & Local Variance (LV)   & 0.174 & 0.222 & 0.222 & \textbf{0.114} & 0.191 & 0.113 \\
                        & Loss Disparity (LD)   & 0.447 & 0.442 & 0.484 & \textbf{0.168} & .120 & 0.166 \\
                        & MIA Success Ratio     & 0.560 & 0.640 & \textbf{0.370} & 0.460 & 0.530 & 0.680 \\
                        & DA Leakage Rate       & 0.011 & 0.007 & 0.006 & 0.005 & 0.005 & \textbf{0.003} \\ \midrule
\multirow{5}{*}{MRI}    & Accuracy (PA)         & 0.994 & 0.495 & \textbf{0.998} & 0.992 & 0.992 & 0.996 \\
                        & Local Variance (LV)   & 0.000 & 0.003 & \textbf{0.000} & 0.000 & 0.000 & 0.000 \\
                        & Loss Disparity (LD)   & 0.000 & NaN & \textbf{0.000} & 0.001 & 0.000 & 0.001 \\
                        & MIA Success Ratio     & 0.610 & 0.690 & \textbf{0.420} & 0.490 & 0.580 & 0.750 \\
                        & DA Leakage Rate       & 0.015 & 0.009 & \textbf{0.008} & 0.007 & 0.007 & 0.004 \\ \midrule
\multirow{5}{*}{CCFD}   & Accuracy (PA)         & 0.998 & 0.998 & 0.998 & 0.998 & 0.998 & 0.998 \\
                        & Local Variance (LV)   & 3.75E-07 & 3.75E-07 & 3.75E-07 & 3.75E-07 & 3.75E-07 & 3.75E-07 \\
                        & Loss Disparity (LD)   & 1.20E-05 & 1.19E-05 & 1.20E-05 & 1.20E-05 & 1.20E-05 & 1.20E-05 \\
                        & MIA Success Ratio     & 0.610 & 0.690 & 0.490 & 0.490 & 0.490 & 0.750 \\
                        & DA Leakage Rate       & 5.49E-01 & 5.53E-01 & 5.49E-01 & 5.49E-01 & 5.00E-01 & 5.49E-01 \\ \toprule

 & & \textbf{SMC} & \textbf{LDP+HE} & \textbf{GDP+HE} & \textbf{LDP+SMC} & \textbf{GDP+SMC} & \textbf{HE+SMC} \\ \midrule
\multirow{5}{*}{MNIST}  & Accuracy (PA)         & 0.963 & 0.966 & 0.952 & 0.962 & 0.951 & \textbf{0.967} \\
                        & Local Variance (LV)   & 0.001 & 0.001 & 0.001 & 0.001 & 0.001 & \textbf{0.000} \\
                        & Loss Disparity (LD)   & 0.009 & 0.008 & 0.018 & 0.010 & 0.020 & \textbf{0.007} \\
                        & MIA Success Ratio     & 0.630 & 0.630 & 0.660 & 0.660 & 0.650 & \textbf{0.640} \\
                        & DA Leakage Rate       & 0.001 & 0.010 & 0.006 & 0.001 & 0.005 & \textbf{0.002} \\ \midrule
\multirow{5}{*}{FMNIST} & Accuracy (PA)         & 0.532 & 0.435 & 0.330 & 0.435 & 0.337 & \textbf{0.534} \\
                        & Local Variance (LV)   & 0.115 & 0.164 & 0.203 & 0.164 & 0.215 & \textbf{0.115} \\
                        & Loss Disparity (LD)   & 0.170 & 0.300 & 0.983 & 0.300 & \textbf{0.010} & 0.169 \\
                        & MIA Success Ratio     & 0.670 & 0.730 & 0.720 & 0.710 & 0.720 & \textbf{0.710} \\
                        & DA Leakage Rate       & 0.001 & 0.014 & 0.009 & 0.001 & 0.007 & \textbf{0.003} \\ \midrule
\multirow{5}{*}{MRI}    & Accuracy (PA)         & 0.994 & 0.992 & 0.495 & 0.992 & 0.495 & \textbf{0.996} \\
                        & Local Variance (LV)   & 0.000 & \textbf{0.000} & 0.003 & 0.000 & 0.003 & 0.000 \\
                        & Loss Disparity (LD)   & 0.001 & \textbf{0.001} & NaN & 0.001 & NaN & NaN \\
                        & MIA Success Ratio     & 0.740 & 0.790 & \textbf{0.780} & 0.780 & 0.790 & 0.790 \\
                        & DA Leakage Rate       & 0.001 & 0.019 & 0.012 & 0.002 & 0.009 & \textbf{0.005} \\ \midrule
\multirow{5}{*}{CCFD}   & Accuracy (PA)         & 0.998 & 0.998 & 0.998 & 0.998 & 0.998 & 0.998 \\
                        & Local Variance (LV)   & 3.75E-07 & 3.75E-07 & 3.75E-07 & 3.75E-07 & 3.75E-07 & 3.75E-07 \\
                        & Loss Disparity (LD)   & 1.20E-05 & 1.20E-05 & 1.20E-05 & 1.20E-05 & 1.20E-05 & 1.20E-05 \\
                        & MIA Success Ratio     & 0.740 & 0.790 & 0.780 & 0.780 & 0.790 & 0.790 \\
                        & DA Leakage Rate       & 5.49E-01 & 5.49E-01 & 5.49E-01 & 5.49E-01 & 5.49E-01 & 5.49E-01 \\ \bottomrule
\end{tabular}
\label{tab:pp-fl-performance-iid}

(Note: In Loss Disparity, the dominance of added noise over the gradients can cause updates to diverge, potentially resulting in NaN values.)
\end{table*}

\begin{table*}[!t]
\centering
\caption{\sc Performance Comparison of Privacy-Preserving FL Algorithms under non-IID Datasets}
\label{tab:pp-fl-performance-non-iid}
\begin{tabular}{@{}llccccccc@{}}
\toprule
\textbf{Dataset} & \textbf{Metric} & \textbf{LDP} & \textbf{GDP} & \textbf{K-Anonymity} & \textbf{L-Anonymity} & \textbf{T-Closeness} & \textbf{HE} \\ \midrule
\multirow{5}{*}{MNIST}  
& Accuracy (PA)         & 0.900 & 0.898 & 0.742 & 0.898 & 0.895 & 0.895 \\
& Local Variance (LV)   & 0.005 & 0.007 & \textbf{0.001} & 0.007 & 0.007 & 0.007 \\
& Loss Disparity (LD)   & 0.068 & \textbf{0.063} & 0.075 & 0.081 & 0.085 & 0.078 \\
& MIA Success Ratio     & 0.530 & 0.610 & \textbf{0.360} & 0.430 & 0.510 & 0.630 \\
& DA Leakage Rate       & 0.007 & 0.004 & 0.004 & 0.004 & 0.003 & \textbf{0.002} \\ \midrule
\multirow{5}{*}{FMNIST} 
& Accuracy (PA)         & 0.280 & 0.283 & \textbf{0.440} & 0.280 & 0.280 & 0.280 \\
& Local Variance (LV)   & 0.202 & 0.200 & 0.246 & 0.202 & \textbf{0.200} & 0.202 \\
& Loss Disparity (LD)   & 0.533 & 0.443 & 0.524 & 0.641 & 0.446 & \textbf{0.636} \\
& MIA Success Ratio     & 0.600 & 0.670 & 0.410 & 0.490 & 0.550 & \textbf{0.710} \\
& DA Leakage Rate       & 0.014 & 0.008 & 0.007 & 0.006 & 0.006 & \textbf{0.004} \\ \midrule
\multirow{5}{*}{MRI}    
& Accuracy (PA)         & 0.989 & 0.742 & 0.973 & 0.980 & \textbf{0.981} & 0.978 \\
& Local Variance (LV)   & 0.001 & 0.110 & 0.001 & \textbf{0.001} & 0.000 & 0.001 \\
& Loss Disparity (LD)   & 0.002 & NaN & 0.008 & 0.003 & 0.003 & \textbf{0.001} \\
& MIA Success Ratio     & 0.640 & 0.740 & 0.440 & 0.530 & 0.600 & \textbf{0.780} \\
& DA Leakage Rate       & 0.019 & 0.011 & 0.010 & 0.009 & 0.008 & \textbf{0.005} \\ \midrule
\multirow{5}{*}{CCFD}   & Accuracy (PA)         & 0.998 & 0.998 & 0.998 & 0.998 & 0.998 & 0.998 \\
                        & Local Variance (LV)   & 7.89E-06 & 3.75E-07 & 7.89E-06 & 7.89E-06 & 7.89E-06 & 7.89E-06 \\
                        & Loss Disparity (LD)   & 1.74E-04 & 1.19E-05 & 1.74E-04 & 1.74E-04 & 1.20E-05 & 1.74E-04 \\
                        & MIA Success Ratio     & 0.810 & 0.820 & 0.780 & 0.780 & 0.780 & 0.850 \\
                        & DA Leakage Rate       & 0.662 & 0.670 & 0.662 & 0.662 & 0.662 & 0.662 \\ \toprule

 & & \textbf{SMC} & \textbf{LDP+HE} & \textbf{GDP+HE} & \textbf{LDP+SMC} & \textbf{GDP+SMC} & \textbf{HE+SMC} \\ \midrule
\multirow{5}{*}{MNIST}  
& Accuracy (PA)         & 0.896 & 0.901 & 0.799 & 0.900 & 0.562 & \textbf{0.897} \\
& Local Variance (LV)   & 0.007 & \textbf{0.006} & 0.018 & 0.006 & 0.008 & 0.007 \\
& Loss Disparity (LD)   & 0.076 & \textbf{0.074} & 0.153 & 0.068 & 0.651 & 0.075 \\
& MIA Success Ratio     & 0.670 & 0.690 & \textbf{0.670} & 0.700 & 0.680 & 0.680 \\
& DA Leakage Rate       & 0.001 & 0.010 & 0.006 & \textbf{0.001} & 0.004 & 0.002 \\ \midrule
\multirow{5}{*}{FMNIST} 
& Accuracy (PA)         & 0.280 & 0.280 & \textbf{0.282} & 0.280 & \textbf{0.282} & 0.280 \\
& Local Variance (LV)   & 0.202 & 0.202 & \textbf{0.200} & 0.202 & \textbf{0.200} & 0.202 \\
& Loss Disparity (LD)   & 0.637 & 0.533 & 0.484 & 0.533 & \textbf{0.431} & 0.517 \\
& MIA Success Ratio     & 0.710 & 0.750 & 0.770 & 0.750 & 0.750 & \textbf{0.760} \\
& DA Leakage Rate       & 0.001 & 0.017 & 0.011 & \textbf{0.001} & 0.008 & 0.004 \\ \midrule
\multirow{5}{*}{MRI}    
& Accuracy (PA)         & 0.981 & 0.981 & 0.743 & 0.979 & 0.743 & \textbf{0.981} \\
& Local Variance (LV)   & 0.001 & \textbf{0.000} & 0.110 & 0.001 & 0.110 & 0.001 \\
& Loss Disparity (LD)   & 0.002 & \textbf{0.003} & NaN & 0.003 & NaN & NaN \\
& MIA Success Ratio     & 0.780 & 0.810 & \textbf{0.780} & 0.820 & 0.800 & 0.800 \\
& DA Leakage Rate       & 0.001 & 0.023 & 0.012 & 0.002 & 0.009 & \textbf{0.005} \\ \midrule
\multirow{5}{*}{CCFD}   & Accuracy (PA)         & 0.998 & 0.998 & 0.998 & 0.998 & 0.998 & 0.998 \\
                        & Local Variance (LV)   & 7.89E-06 & 7.89E-06 & 7.89E-06 & 7.89E-06 & 7.89E-06 & 7.89E-06 \\
                        & Loss Disparity (LD)   & 1.74E-04 & 1.74E-04 & 1.74E-04 & 1.74E-04 & 1.74E-04 & 1.74E-04 \\
                        & MIA Success Ratio     & 0.850 & 0.850 & 0.850 & 0.850 & 0.850 & 0.850 \\
                        & DA Leakage Rate       & 0.662 & 0.662 & 0.662 & 0.662 & 0.662 & 0.662 \\ \bottomrule
\end{tabular}

(Note: In Loss Disparity, the dominance of added noise over the gradients can cause updates to diverge, potentially resulting in NaN values.)
\end{table*}

\section{Trade-Off Results \& Analysis} \label{sec:results-analyses}

%This section evaluates the proposed framework under varying fairness and privacy conditions across three datasets in Section~\ref{sec:exp-setup}. A detailed analysis of the results follows.

\begin{figure*}[ht]
    \centering
    \includegraphics[width=\textwidth]{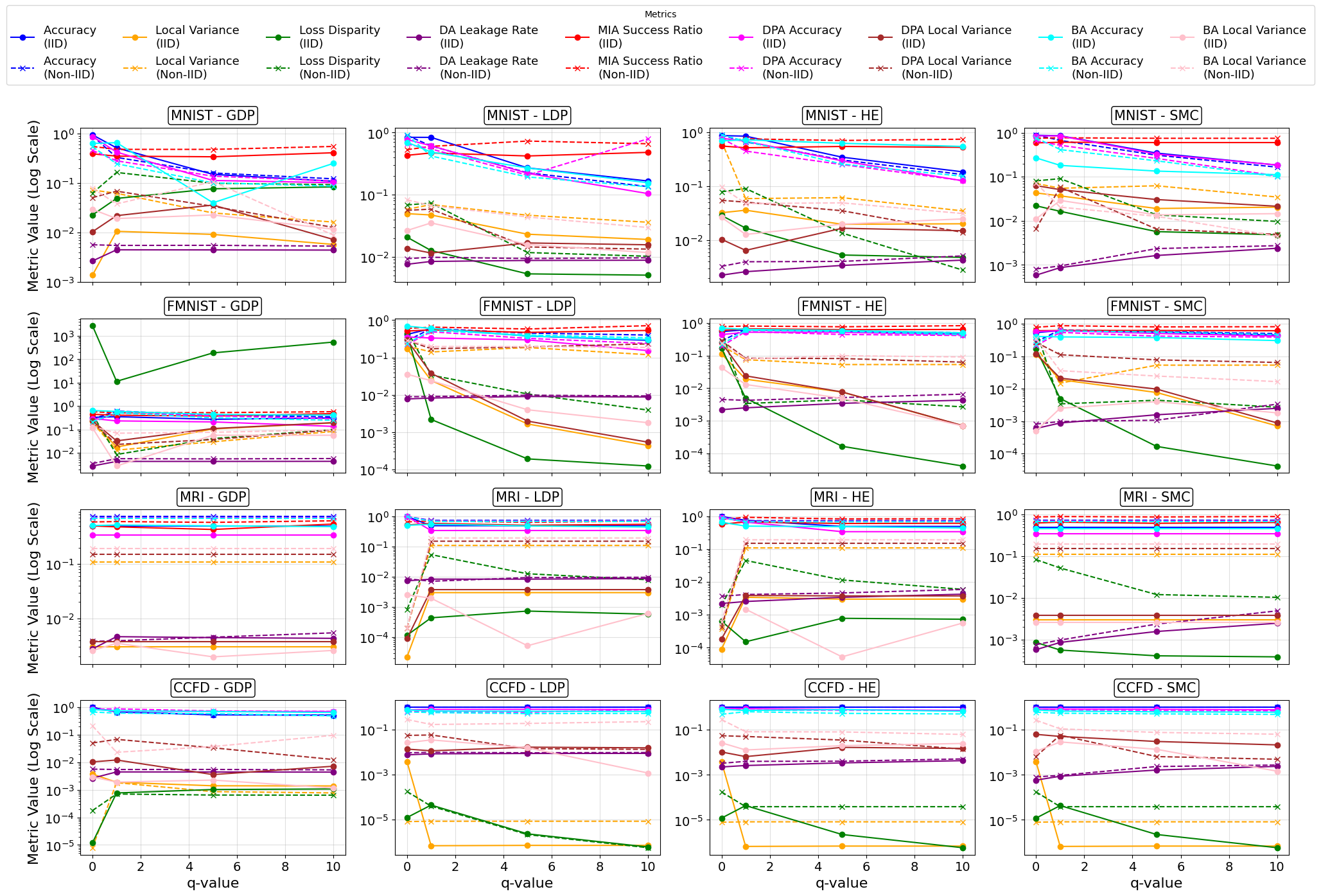} 
    \caption{Impact of privacy-preserving FL algorithms (GDP, LDP, HE, and SMC) across four datasets under IID and non-IID conditions, evaluated for different \( q \)-values (0, 1, 5, 10).}
    \label{fig:graph1} % Replace with your label for referencing
\end{figure*}

\subsection{Effect of Varying Fairness on Privacy Risk}

We analyze how varying fairness levels impact privacy risks in q-FedAvg using DP, HE (CKKS variant), and SMC (Shamir’s Secret Sharing). The fairness parameter \(q\) (\(0, 1, 5, 10\)) is adjusted to assess fairness-privacy trade-offs across four datasets under IID and non-IID settings. Privacy is evaluated using MIA success ratio and DA leakage rate in white-box settings, with 10\% of data exposed for MIAs and gradient changes analyzed over 800 interactions for DA leakage. Prediction accuracy assesses utility.

\textbf{Effect of Varying Fairness under DP.} Using GDP and LDP, q-FedAvg is evaluated across four datasets under IID and non-IID settings with a fixed privacy budget (\(\epsilon = 8\)). As in Figure~\ref{fig:graph1}, increasing fairness levels (\(q\)) reduce accuracy, with a stronger decline in non-IID settings due to higher data heterogeneity. In IID settings, accuracy decreases more gradually but remains notable, reflecting the fairness-utility trade-off.
Privacy risks, measured by DA leakage rates and MIA success ratios, increase with \(q\), as DP noise, while enhancing fairness, amplifies vulnerabilities. DA leakage rates rise moderately under GDP and LDP, while MIA success ratios fluctuate more in non-IID settings. LDP offers better privacy protection than GDP, as localized noise more effectively obfuscates client contributions.

Under adversarial attacks, backdoor accuracy declines with fairness, while non-IID settings show higher variance in local performance. Data poisoning accuracy varies across IID and non-IID settings, indicating dataset-dependent robustness trade-offs. LDP exhibits stronger privacy robustness than GDP due to its localized noise application.

\begin{figure*}[!t]
    \centering
    \includegraphics[width=0.99\textwidth]{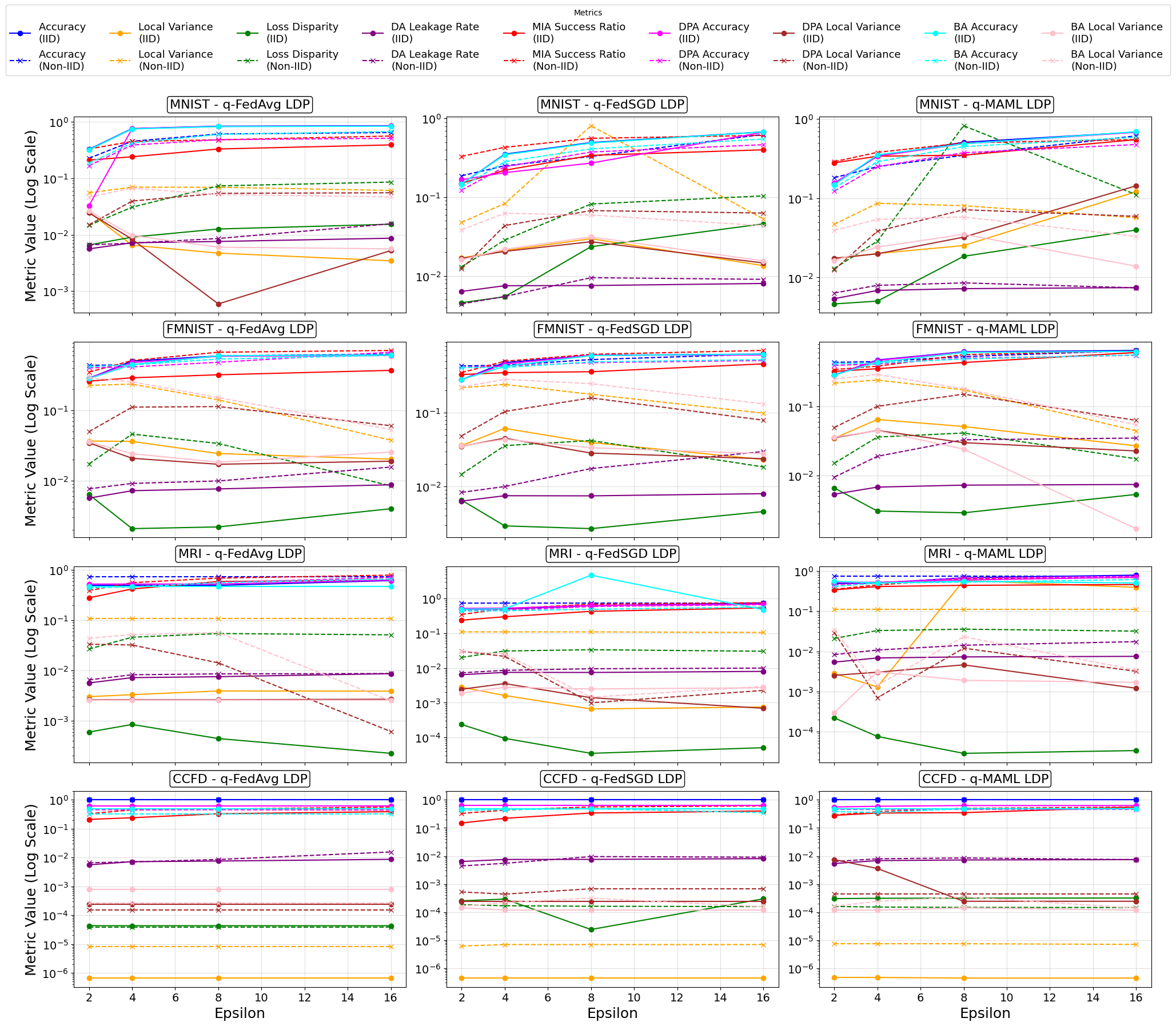}
    \caption{Evaluation of fair FL algorithms with DP: Each row represents a dataset under IID and non-IID conditions, while each column corresponds to a fair FL algorithm (q-FedAvg, q-FedSGD, q-MAML).}
    \label{fig:graph2a} % Replace with your label for referencing
\end{figure*}

\begin{figure*}[!t]
    \centering
    \includegraphics[width=0.99\textwidth]{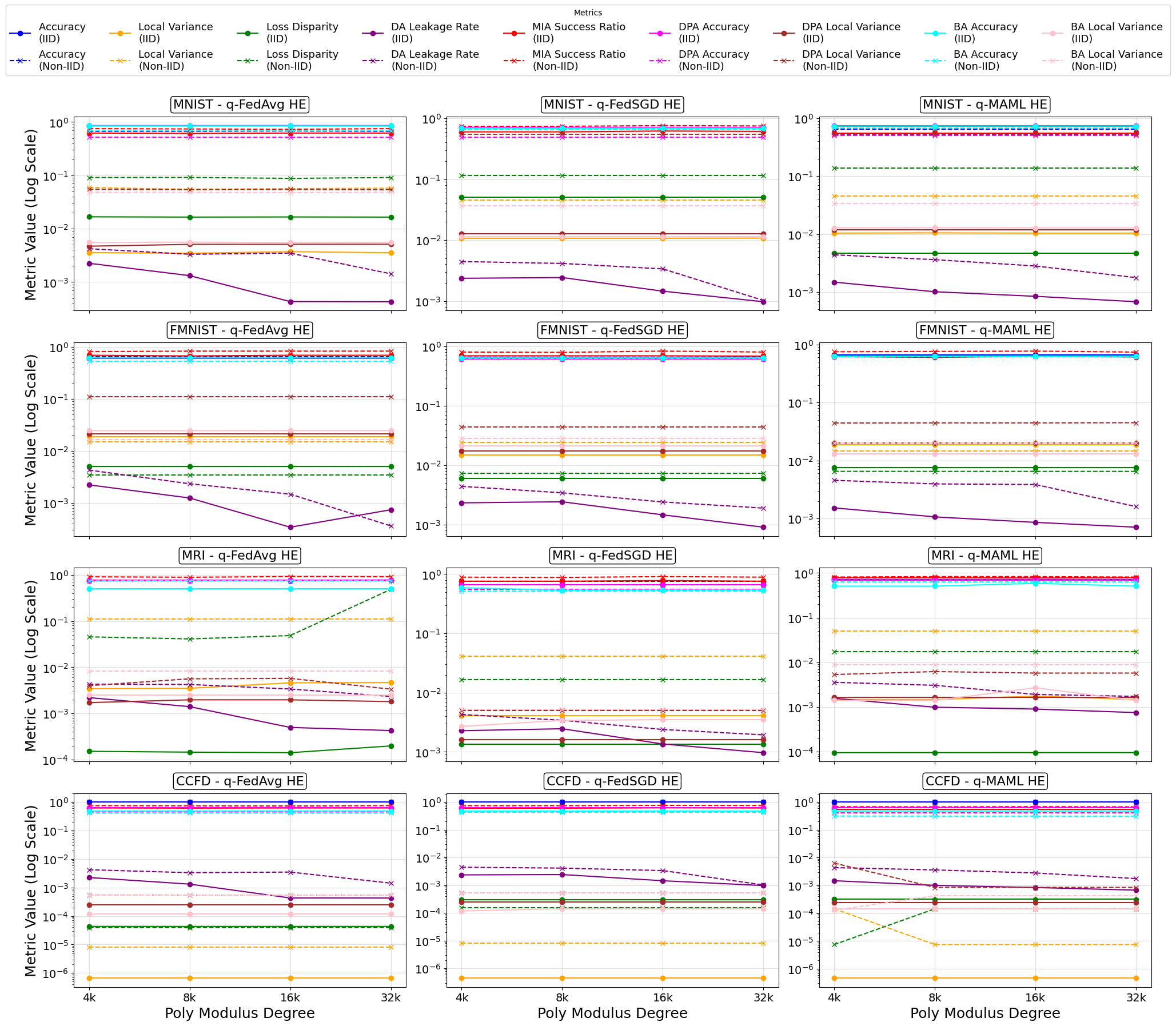} 
    \caption{Impact of fair FL algorithms with HE: Each row represents a dataset under IID and non-IID conditions, while each column corresponds to a fair FL algorithm (q-FedAvg, q-FedSGD, q-MAML).}
    \label{fig:graph2b} % Replace with your label for referencing
\end{figure*}

\textbf{Effect of Varying Fairness under HE.} Under HE, q-FedAvg is evaluated with a polynomial modulus degree of 4096 for strong privacy. We assess the impact of fairness levels on the MIA success ratio and DA leakage rate.

As shown in Figure~\ref{fig:graph1}, HE maintains greater stability in privacy metrics than DP as \(q\) increases. DA leakage and MIA success ratios rise slightly, showing HE preserves privacy despite fairness adjustments. Accuracy declines with \(q\), in non-IID settings, but less than under DP, indicating fewer constraints on model updates and better utility retention.

Fairness metrics, such as local variance and loss disparity, consistently decrease with higher \(q\), particularly in non-IID settings, suggesting HE balances fairness and privacy while reducing performance disparities. Under adversarial attacks, HE shows clear IID and non-IID differences. In non-IID settings, backdoor attack accuracy is lower with increased local variance, while data poisoning metrics distinguish IID from non-IID cases. HE retains better utility while balancing fairness and privacy, particularly in Fashion-MNIST.

\textbf{Effect of Varying Fairness under SMC.} For SMC results in Figure~\ref{fig:graph1} (last column), gradients are securely shared with 7 fixed shares and an AES key size of 128. The q-FedAvg optimizer is evaluated across varying \(q\) levels for four datasets under IID and non-IID conditions.  SMC maintains stable privacy and fairness metrics as \(q\) increases, with minimal rises in DA leakage and MIA success ratios, reflecting strong privacy guarantees. Unlike DP and HE, SMC achieves the lowest loss disparity, effectively managing fairness adjustments while preserving privacy.  Accuracy under SMC follows HE, decreasing with \(q\) but retaining better utility than DP, particularly in non-IID settings. LV steadily declines, reinforcing SMC’s balance of fairness and privacy. In MRI, SMC outperforms DP and HE stably, ideal for privacy-critical applications with fairness constraints.

Backdoor and data poisoning attack metrics remain stable under SMC, which consistently achieves the lowest loss disparity. This is evident in MRI, where SMC demonstrates superior stability. The results show secure gradient sharing mitigates adverse effects of fairness adjustments, preserving utility and robustness against adversarial manipulations.

\subsection{Effect of Varying Privacy Levels on Fairness} \label{subsec:varying-privacy-on-fairness}

We explore the effect of varying PP parameters on fairness in FL by evaluating DP and HE under q-FedAvg with \(q=1\).

\textbf{Effect of Varying Privacy Levels in DP on Fairness.} We vary the privacy budget \(\epsilon\) for LDP while keeping other parameters fixed. Across the four datasets, fairness improves with higher \(\epsilon\), especially in IID settings where local variance decreases, reducing model contribution disparities. In MNIST, increased \(\epsilon\) mitigates LDP noise effects, enhancing fairness.  In non-IID settings, fairness gains are slower, while LD remains sensitive to imbalances, with smaller reductions as \(\epsilon\) increases. DA leakage rates rise moderately at higher \(\epsilon\), challenging the privacy-fairness balance.  DPA effects on accuracy and local variance are inconsistent. While higher \(\epsilon\) reduces noise impact, adversarial sample distributions under LDP cause fairness fluctuations. DPA can lower accuracy and increase local variance, varying with LDP's sensitivity to noise and non-IID heterogeneity.  Higher \(\epsilon\) improves fairness while preserving privacy in IID settings. In non-IID conditions, careful calibration ensures equitable client contributions without privacy loss. Backdoor and data poisoning fairness levels decline slightly with higher \(\epsilon\), stressing the need for adversarial resilience in fairness-critical applications like CCFD and AD MRI.

\textbf{Effect of Varying Privacy Levels in HE on Fairness.} The effect of HE on fairness is analyzed by varying the polynomial modulus degree, balancing privacy and efficiency. Figure~\ref{fig:graph2b} shows that higher modulus degrees improve local variance and loss disparity, particularly in non-IID settings, due to HE’s secure aggregation ensuring consistent updates.  In IID settings, fairness stabilizes quickly. In non-IID conditions, FMNIST and CCFD show reduced loss disparity, though larger modulus degrees increase computational overhead. HE maintains stable fairness under adversarial poisoning, ensuring predictable performance.  Unlike DP, HE preserves privacy with minimal impact on DA leakage and MIA success, even as fairness improves. Fairness metrics under adversarial attacks remain stable, highlighting HE’s robustness in fairness-sensitive applications.

\section{Performance Analysis of PP FL Algorithms under IID and non-IID Datasets} \label{appen:pp-fl}

{\bf Under IID Datasets}: Table~\ref{tab:pp-fl-performance-iid} highlights the performance of various privacy-preserving mechanisms across four datasets under IID conditions. In the MNIST dataset, \( K \)-Anonymity demonstrates the best performance in terms of accuracy, achieving the highest value, while HE+SMC excels in minimizing local variance, indicating its ability to maintain uniform client performance. For loss disparity, $L$-Anonymity and HE+SMC perform exceptionally well by reducing disparities effectively. \( K \)-Anonymity provides the strongest privacy protection, with the lowest MIA success ratio. HE ensures robust data privacy with the lowest DA leakage rate.  

For the FMNIST dataset, \( T \)-Closeness achieves the highest accuracy, closely followed by HE. In terms of fairness, HE and SMC demonstrate the lowest local variance, ensuring consistency across client contributions. HE also leads in minimizing loss disparity, slightly outperforming \( T \)-Closeness. Privacy metrics reveal that \( K \)-Anonymity is the most effective against membership inference attacks with the lowest success ratio, while HE achieves the lowest DA leakage rate, highlighting its superior privacy-preserving capabilities.  

In the MRI dataset, \( K \)-Anonymity achieves the highest accuracy, showcasing its effectiveness in handling this dataset. Both \( K \)-Anonymity and GDP+HE minimize local variance, reflecting strong fairness. Loss disparity is effectively addressed by \( K \)-Anonymity, demonstrating resilience against membership inference attacks with the lowest success ratio. HE, on the other hand, ensures robust privacy preservation by achieving the lowest DA leakage rate.

Overall, \( K \)-Anonymity emerges as a strong contender for achieving high accuracy and fairness across datasets, while HE and HE+SMC consistently excel in privacy preservation and fairness metrics. These findings underscore the importance of selecting algorithms that align with the specific characteristics of the dataset and the desired trade-offs between privacy, fairness, and accuracy.

{\bf Under Non-IID Datasets}: Table~\ref{tab:pp-fl-performance-non-iid} illustrates the performance of privacy-preserving mechanisms across four datasets under non-IID conditions. In the MNIST dataset, \( K \)-Anonymity achieves the highest accuracy, demonstrating its robustness in non-IID settings. For fairness, \( K \)-Anonymity also achieves the lowest local variance, indicating its ability to maintain uniform client performance. In loss disparity, GDP performs the best by minimizing disparities effectively. Regarding privacy, \( K \)-Anonymity stands out with the lowest MIA success ratio, while HE ensures the strongest data privacy with the lowest DA leakage rate.

For the FMNIST dataset, \( K \)-Anonymity achieves the highest accuracy, showcasing its adaptability to this dataset. In terms of fairness, both GDP and \( T \)-Closeness exhibit the lowest local variance, highlighting their ability to maintain consistent model updates across clients. HE leads in minimizing loss disparity, slightly outperforming other algorithms. Privacy metrics reveal that \( K \)-Anonymity achieves the lowest MIA success ratio.  HE consistently demonstrates robust privacy protection with the lowest DA leakage rate.

In the MRI dataset, \( T \)-Closeness delivers the highest accuracy, emphasizing its effectiveness in addressing data heterogeneity in this challenging dataset. \( K \)-Anonymity achieves the lowest local variance, ensuring fairness by reducing performance disparities among clients. HE demonstrates the lowest loss disparity, highlighting its ability to handle varying data distributions effectively. For privacy preservation, \( K \)-Anonymity achieves the lowest MIA success ratio, while HE exhibits strong privacy performance with the lowest DA leakage rate.

Across all datasets, \( K \)-Anonymity emerges as a leading contender for achieving high accuracy and fairness, while HE consistently excels in privacy preservation metrics. These findings show the importance of selecting privacy-preserving FL algorithms based on the specific trade-offs required by the dataset characteristics and performance objectives.

Comparing the performance of privacy-preserving FL mechanisms under IID and non-IID datasets reveals notable distinctions in algorithm effectiveness. Under IID conditions (Table~\ref{tab:pp-fl-performance-iid}), \( K \)-Anonymity consistently achieves high accuracy across datasets, while HE and HE+SMC demonstrate exceptional capabilities in minimizing local variance and ensuring fairness. In particular, HE emerges as a leader in privacy preservation, achieving the lowest DA leakage rate across multiple datasets. On the other hand, non-IID scenarios (Table~\ref{tab:pp-fl-performance-non-iid}) highlight the adaptability of \( K \)-Anonymity, which continues to excel in accuracy and fairness metrics despite the challenges of heterogeneous data distributions. HE remains a top performer in privacy metrics, maintaining the lowest DA leakage rates under non-IID conditions.

The MRI dataset underscores these distinctions more prominently due to its inherent data heterogeneity. While \( K \)-Anonymity delivers high accuracy and minimizes local variance in both settings, HE effectively addresses loss disparity and ensures robust privacy in both IID and non-IID scenarios. However, \( T \)-Closeness outperforms other methods in accuracy for MRI under non-IID conditions, highlighting the need to tailor algorithm selection to dataset characteristics and performance objectives. These findings emphasize that while certain algorithms like \( K \)-Anonymity and HE are versatile across data distributions, the interplay between privacy, fairness, and accuracy necessitates a context-aware approach for selecting the most suitable algorithm.

\section{Performance Comparison of Fair FL Algorithms under IID and non-IID Datasets} \label{append:fair-fl-comparison}

{\bf Under IID Datasets}: Table~\ref{tab:fair-fl-performance-iid} compares the performance of fair FL algorithms under IID conditions across four datasets.  Table~\ref{tab:fair-fl-performance-iid} presents the performance of Fair FL algorithms under IID datasets across four datasets: MNIST, FMNIST, MRI and CCFD. For MNIST, \( \text{q-FedAvg} \) achieves the highest accuracy, indicating its strength in maintaining overall model performance. However, \( \text{Ditto} \) stands out in fairness-related metrics, achieving the lowest local variance and loss disparity, showcasing its ability to ensure uniformity across clients. Furthermore, \( \text{Ditto} \) also achieves the lowest DA leakage rate, highlighting its superior privacy preservation. In terms of adversarial privacy accuracy (DPA Accuracy), \( \text{q-FedAvg} \) demonstrates the best performance, and it also minimizes DPA local variance, reinforcing its robustness in adversarial scenarios.

For FMNIST, \( \text{Ditto} \) dominates in multiple metrics. It achieves the highest accuracy, the lowest local variance, and the lowest loss disparity, underlining its robustness in both accuracy and fairness. Additionally, \( \text{Ditto} \) achieves the lowest DA leakage rate, further emphasizing its strong privacy preservation capabilities. In adversarial privacy metrics, \( \text{Ditto} \) maintains superiority by achieving the highest DPA Accuracy and the lowest DPA local variance, making it a comprehensive performer under FMNIST.

In the MRI dataset, \( \text{Ditto} \) emerges as the best performer in accuracy, DPA Accuracy, and DPA local variance, showcasing its overall adaptability and strength. For fairness, \( \text{AFL} \) performs exceptionally well, achieving the lowest local variance and loss disparity. Moreover, \( \text{AFL} \) also attains the lowest DA leakage rate, highlighting its ability to balance fairness and privacy effectively. However, in terms of adversarial privacy success, \( \text{q-MAML} \) shows a significant edge with the lowest MIA success ratio.

Across datasets, \( \text{Ditto} \) consistently performs well in accuracy, fairness, and privacy metrics, making it a versatile option. \( \text{AFL} \) also demonstrates strong performance in privacy preservation and fairness for MRI. These results underscore the importance of selecting FL algorithms tailored to the specific requirements of the dataset and the desired trade-offs between accuracy, fairness, and privacy.

{\bf Under Non-IID Datasets}: Table~\ref{tab:fair-fl-performance-non-iid} compares the performance of fair FL algorithms under non-IID conditions across four datasets. Table~\ref{tab:fair-fl-performance-non-iid} evaluates the performance of Fair FL algorithms under non-IID datasets across MNIST, FMNIST, and MRI datasets. For the MNIST dataset, AFL demonstrates the best accuracy, highlighting its capability to adapt to non-IID data distributions effectively. It also achieves the lowest local variance, indicating consistent performance across clients. In loss disparity, AFL achieves the lowest value, showcasing its ability to minimize differences in model updates among clients. AFL further excels in adversarial privacy metrics, achieving the highest DPA accuracy and the lowest DPA local variance, reinforcing its strength in adversarial scenarios. Regarding DA leakage rate, AFL and Ditto share the lowest value, emphasizing their robust privacy preservation.

In the FMNIST dataset, Ditto achieves the highest accuracy, showcasing its adaptability to challenging non-IID conditions. For fairness, both q-SGD and q-FedAvg attain the lowest local variance, ensuring uniform model performance across clients. Regarding loss disparity, Ditto achieves the lowest value, reflecting its ability to mitigate disparities in model updates effectively. Ditto also stands out in privacy metrics, achieving the lowest DA leakage rate and the highest DPA accuracy, underscoring its comprehensive privacy-preserving capabilities.

For the MRI dataset, Ditto again demonstrates the best accuracy, affirming its robustness under highly heterogeneous data distributions. It achieves the lowest local variance, highlighting its fairness in maintaining consistent client performance. Loss disparity is minimized by AFL, demonstrating its ability to handle disparate client data effectively. Regarding privacy metrics, AFL achieves the lowest DA leakage rate, while q-MAML attains the highest DPA accuracy, demonstrating strong privacy and adversarial robustness. AFL also achieves the lowest DPA local variance, showcasing its ability to maintain consistent privacy across clients.

Overall, AFL emerges as a leading contender for maintaining fairness and privacy in non-IID scenarios, while Ditto consistently achieves high accuracy and strong privacy-preserving performance across datasets. These results underscore the importance of tailoring algorithm selection to the specific trade-offs required for non-IID data distributions.

The comparison of Fair FL algorithms under IID and non-IID datasets, as shown in Tables~\ref{tab:fair-fl-performance-iid} and~\ref{tab:fair-fl-performance-non-iid}, highlights distinct patterns in algorithm performance across MNIST, FMNIST, and MRI datasets. Under IID conditions, \( \text{Ditto} \) consistently excels in fairness and privacy metrics across all datasets, achieving the lowest local variance, loss disparity, and DA leakage rate. It also demonstrates robust accuracy, particularly for FMNIST and MRI, making it versatile in IID scenarios. In contrast, \( \text{AFL} \) performs well in MRI by achieving the lowest local variance, loss disparity, and DA leakage rate, balancing fairness and privacy. \( \text{q-FedAvg} \) leads in adversarial privacy metrics (DPA Accuracy and DPA local variance) for MNIST, reinforcing its strength in adversarial scenarios.

Under non-IID conditions, the results shift significantly. \( \text{AFL} \) dominates for MNIST, achieving the highest accuracy and excelling in fairness metrics such as local variance and loss disparity. Similarly, \( \text{Ditto} \) is a robust performer for FMNIST and MRI, achieving the highest accuracy and consistently minimizing DA leakage rate and loss disparity. While \( \text{q-SGD} \) and \( \text{q-FedAvg} \) perform competitively in fairness metrics for FMNIST, \( \text{q-MAML} \) shows strong adversarial privacy protection for MRI, achieving the highest DPA accuracy.

Overall, while \( \text{Ditto} \) maintains consistent performance across both IID and non-IID scenarios, \( \text{AFL} \) demonstrates adaptability to non-IID data, particularly in MNIST and MRI datasets. These findings underscore the need to select algorithms tailored to the data distribution and specific trade-offs between accuracy, fairness, and privacy required for the application context.

\begin{table*}[!t]
\renewcommand{\arraystretch}{1.1} 
\centering
\caption{\sc Performance Comparison of Fair FL Algorithms under IID Datasets} 
\label{tab:fair-fl-performance-iid}
\begin{tabular}{@{}llccccc@{}}
\toprule
\textbf{Dataset} & \textbf{Metric} & \textbf{q-FedAvg} & \textbf{q-SGD} & \textbf{q-MAML} & \textbf{AFL} & \textbf{Ditto} \\
\midrule
\multirow{5}{*}{MNIST} 
& Accuracy (PA)         & \textbf{0.855} & 0.707 & 0.722 & 0.739 & 0.700 \\
& Local Variance (LV)   & 0.004 & 0.011 & 0.010 & 0.012 & \textbf{0.003} \\
& Loss Disparity (LD)   & 0.017 & 0.051 & 0.046 & 0.050 & \textbf{0.016} \\
& MIA Success Ratio     & 0.640 & 0.630 & \textbf{0.520} & 0.680 & 0.550 \\
& DA Leakage Rate       & 0.007 & 0.008 & 0.007 & 0.005 & \textbf{0.004} \\
\midrule

\multirow{5}{*}{FMNIST} 
& Accuracy (PA)         & 0.617 & 0.641 & 0.654 & 0.627 & \textbf{1.000} \\
& Local Variance (LV)   & 0.019 & \textbf{0.015} & 0.019 & 0.031 & \textbf{0.000} \\
& Loss Disparity (LD)   & \textbf{0.005} & 0.006 & 0.008 & 0.061 & \textbf{0.000} \\
& MIA Success Ratio     & 0.670 & 0.690 & 0.590 & \textbf{0.600} & 0.740 \\
& DA Leakage Rate       & 0.009 & 0.009 & 0.007 & 0.007 & \textbf{0.007} \\
\midrule

\multirow{5}{*}{MRI} 
& Accuracy (PA)         & 0.495 & 0.495 & 0.505 & 0.718 & \textbf{0.790} \\
& Local Variance (LV)   & 0.003 & 0.003 & 0.003 & \textbf{0.002} & 0.003 \\
& Loss Disparity (LD)   & 0.001 & 0.002 & 0.000 & \textbf{0.000} & 0.157 \\
& MIA Success Ratio     & 0.770 & 0.750 & 0.620 & \textbf{0.780} & 0.660 \\
& DA Leakage Rate       & 0.013 & 0.016 & 0.010 & \textbf{0.008} & 0.073 \\
\midrule

\multirow{5}{*}{CCFD} 
& Accuracy (PA)         & 0.998 & 0.998 & 0.998 & 0.998 & \textbf{0.997} \\
& Local Variance (LV)   & 6.61E-07 & 4.44E-07 & 4.63E-07 & 4.44E-07 & \textbf{1.09E-06} \\
& Loss Disparity (LD)   & 4.32E-05 & 3.06E-04 & 3.20E-04 & 1.35E-04 & \textbf{1.02E-04} \\
& MIA Success Ratio     & 0.87 & 0.86 & 0.85 & 0.79 & \textbf{0.79} \\
& DA Leakage Rate       & 0.00786 & 0.00898 & 0.00831 & 0.00835 & \textbf{0.00789} \\
\bottomrule
\end{tabular}
\end{table*}

\begin{table*}[!t]
\renewcommand{\arraystretch}{1.1} 
\centering
\caption{\sc Performance Comparison of Fair FL Algorithms under non-IID Datasets} 
\label{tab:fair-fl-performance-non-iid}
\begin{tabular}{@{}llccccc@{}}
\toprule
\textbf{Dataset} & \textbf{Metric} & \textbf{q-FedAvg} & \textbf{q-SGD} & \textbf{q-MAML} & \textbf{AFL} & \textbf{Ditto} \\
\midrule
\multirow{5}{*}{MNIST} 
& Accuracy (PA)         & 0.670 & 0.679 & 0.649 & \textbf{0.792} & 0.740 \\
& Local Variance (LV)   & 0.055 & 0.046 & 0.046 & \textbf{0.009} & 0.144 \\
& Loss Disparity (LD)   & 0.091 & 0.116 & 0.138 & \textbf{0.762} & 0.919 \\
& MIA Success Ratio     & 0.670 & 0.660 & 0.550 & \textbf{0.690} & 0.590 \\
& DA Leakage Rate       & 0.009 & 0.009 & 0.007 & \textbf{0.006} & 0.006 \\
\midrule

\multirow{5}{*}{FMNIST} 
& Accuracy (PA)         & 0.643 & 0.668 & 0.658 & 0.648 & \textbf{0.832} \\
& Local Variance (LV)   & 0.015 & 0.025 & \textbf{0.015} & 0.097 & 0.104 \\
& Loss Disparity (LD)   & 0.003 & 0.007 & 0.007 & 0.995 & \textbf{0.202} \\
& MIA Success Ratio     & 0.740 & 0.720 & 0.590 & \textbf{0.790} & 0.630 \\
& DA Leakage Rate       & 0.009 & 0.010 & 0.008 & 0.008 & \textbf{0.007} \\
\midrule

\multirow{5}{*}{MRI} 
& Accuracy (PA)         & 0.742 & 0.753 & 0.788 & 0.777 & \textbf{0.953} \\
& Local Variance (LV)   & 0.110 & 0.042 & 0.059 & 0.072 & \textbf{0.003} \\
& Loss Disparity (LD)   & 0.052 & 0.017 & 0.023 & \textbf{0.001} & 0.127 \\
& MIA Success Ratio     & \textbf{0.790} & \textbf{0.790} & 0.660 & 0.750 & 0.700 \\
& DA Leakage Rate       & 0.017 & 0.023 & 0.009 & \textbf{0.009} & 0.075 \\
\midrule

\multirow{5}{*}{CCFD} 
& Accuracy (PA)         & 0.999 & 0.999 & 0.999 & 0.999 & \textbf{0.997} \\
& Local Variance (LV)   & 8.19E-06 & 8.10E-06 & 7.50E-06 & 7.71E-06 & \textbf{2.23E-05} \\
& Loss Disparity (LD)   & 3.86E-05 & 1.54E-04 & 1.45E-04 & 8.03E-05 & \textbf{1.13E-04} \\
& MIA Success Ratio     & 0.84 & 0.81 & 0.77 & 0.78 & \textbf{0.75} \\
& DA Leakage Rate       & 2.30E-02 & 0.00567 & 0.00908 & 0.00873 & \textbf{0.00352} \\
\bottomrule
\end{tabular}
\end{table*}

\begin{table*}[!t]
\renewcommand{\arraystretch}{1.1} 
\centering
\caption{\sc Lattice Attack (LA) Success Ratios for Different Homomorphic Encryption Variants. 
Higher modulus degrees reduce attack success rates, improving security.}
\label{tab:la_he}
\begin{tabular}{@{}llcccc@{}}
\toprule
\textbf{Variant} & \textbf{Dataset} & \textbf{Degree = 4K} & \textbf{Degree = 8K} & \textbf{Degree = 16K} & \textbf{Degree = 32K} \\
\midrule
\multirow{4}{*}{CKKS}  & MNIST  & 0.1361 & 0.0739 & 0.0654 & 0.0563 \\
                       & FMNIST & 0.1458 & 0.0845 & 0.0778 & 0.0703 \\
                       & MRI AD & 0.1596 & 0.1098 & 0.0942 & 0.8468 \\
                       & CCFD   & 0.2643 & 0.2147 & 0.2009 & 0.1679 \\
\midrule
\multirow{4}{*}{BGV}   & MNIST  & 0.0171 & 0.0169 & 0.0167 & 0.0162 \\
                       & FMNIST & 0.0187 & 0.0177 & 0.0171 & 0.0168 \\
                       & MRI AD & 0.0203 & 0.0201 & 0.0195 & 0.0191 \\
                       & CCFD   & 0.0254 & 0.0245 & 0.0221 & 0.0213 \\
\midrule
\multirow{4}{*}{TFHE}  & MNIST  & 0.0817 & 0.0789 & 0.0739 & 0.0656 \\
                       & FMNIST & 0.0819 & 0.0809 & 0.0799 & 0.0731 \\
                       & MRI AD & 0.0892 & 0.0887 & 0.0869 & 0.0891 \\
                       & CCFD   & 0.0933 & 0.0921 & 0.0905 & 0.0891 \\
\bottomrule
\end{tabular}
\end{table*}

\begin{table*}[!t]
\renewcommand{\arraystretch}{1.1} 
\centering
\caption{\sc Share Reconstruction Attack (SRA) Success Ratios for Different Secure Multi-Party Computation Variants. 
Increasing the number of shares reduces the likelihood of successful reconstruction attacks.}
\label{tab:sra_smc}
\begin{tabular}{@{}llcccc@{}}
\toprule
\textbf{Variant} & \textbf{Dataset} & \textbf{Shares = 3} & \textbf{Shares = 5} & \textbf{Shares = 7} & \textbf{Shares = 9} \\
\midrule
\multirow{4}{*}{CKKS}  & MNIST  & 0.303  & 0.182  & 0.13   & 0.101  \\
                       & FMNIST & 0.315  & 0.223  & 0.164  & 0.159  \\
                       & MRI AD & 0.373  & 0.204  & 0.193  & 0.177  \\
                       & CCFD   & 0.405  & 0.298  & 0.263  & 0.234  \\
\midrule
\multirow{4}{*}{BGV}   & MNIST  & 0.606  & 0.600  & 0.593  & 0.0585 \\
                       & FMNIST & 0.616  & 0.607  & 0.601  & 0.595  \\
                       & MRI AD & 0.797  & 0.789  & 0.773  & 0.766  \\
                       & CCFD   & 0.822  & 0.814  & 0.811  & 0.802  \\
\midrule
\multirow{4}{*}{TFHE}  & MNIST  & 0.00254 & 0.00255 & 0.00254 & 0.00253 \\
                       & FMNIST & 0.00264 & 0.00263 & 0.00260 & 0.00260 \\
                       & MRI AD & 0.00269 & 0.00268 & 0.00268 & 0.00268 \\
                       & CCFD   & 0.00272 & 0.00270 & 0.00270 & 0.00270 \\
\bottomrule
\end{tabular}
\end{table*}

\section{Security Analysis of Homomorphic Encryption and Secure Multi-Party Computation}

Tables~\ref{tab:la_he} and~\ref{tab:sra_smc} present the success ratios of Lattice Attacks (LA) on Homomorphic Encryption (HE) and Share Reconstruction Attacks (SRA) on Secure Multi-Party Computation (SMC), illustrating how encryption parameters and threshold shares influence security.

\subsection{Lattice Attack on Homomorphic Encryption}
Table~\ref{tab:la_he} demonstrates that increasing the polynomial modulus degree significantly reduces the success rate of LA across all datasets. This trend is particularly evident in the CKKS variant, where lower modulus degrees (e.g., 4K) exhibit higher attack success rates, particularly for the CCFD dataset. Comparatively, the BGV and TFHE schemes maintain consistently lower attack success rates across varying degrees, suggesting their inherent robustness to lattice-based attacks. These results highlight the trade-off between computational efficiency and security, where higher degrees offer better protection but introduce greater computational overhead.

\subsection{Share Reconstruction Attack on Secure Multi-Party Computation}
Table~\ref{tab:sra_smc} evaluates the success ratios of SRA across different numbers of threshold shares in SMC. The results indicate that increasing the number of shares significantly reduces the likelihood of successful reconstruction attacks. Notably, the CKKS variant shows a gradual decline in attack success as the number of shares increases from 3 to 9. However, the BGV scheme exhibits relatively higher attack success rates, particularly in MRI AD and CCFD, emphasizing the need for higher threshold shares in privacy-critical applications. TFHE, in contrast, maintains the lowest success rates across all settings, reinforcing its suitability for high-security applications despite potential computational overhead.

These findings underscore the importance of selecting appropriate encryption parameters and threshold shares based on security and performance requirements. While higher modulus degrees and share thresholds improve security, they also introduce trade-offs in computational efficiency and latency, requiring careful consideration in FL deployments.

%%%%%%%%%%%%%%%%%%%%%%%%%%%%%%%%%%%

\section{Case Studies on Fairness and Privacy Risks in Federated Learning}

To analyze the real-world impact of fairness and privacy trade-offs in federated learning (FL), we evaluate two critical application domains: medical diagnosis and financial fraud detection. Our experiments on the Alzheimer's Disease MRI dataset and the Credit Card Fraud Detection dataset reveal distinct performance variations due to privacy-preserving techniques, highlighting their effect on fairness in FL.

\subsection{Privacy-Fairness Trade-offs in Federated Alzheimer's Disease Detection}

\textbf{Impact of Privacy Mechanisms on Model Performance.} Our experiments show that applying {Differential Privacy} with a privacy budget of $\epsilon=1$ results in a {14.8\% drop} in classification accuracy compared to the non-private baseline. Furthermore, clients with smaller MRI datasets exhibit a {21.4\% increase} in false negatives, potentially leading to undiagnosed Alzheimer's cases. The loss in predictive power is due to DP's injected noise, which disproportionately affects data-constrained clients, degrading fairness.

\textbf{Homomorphic Encryption and Secure Multi-Party Computation Trade-offs.} While HE and SMC provide robust encryption, our results show a {32.5\% increase} in model aggregation latency. This affects fairness, as clients with limited resources face delayed updates, causing performance variance across institutions. Well-resourced hospitals benefit from fresher models, while underfunded ones use outdated versions, leading to diagnostic disparities.

\textbf{Fairness Degradation Due to Data Heterogeneity.} Clients with lower-quality MRI scans and imbalanced data suffer most under privacy constraints. With {Differential Privacy}, classification accuracy for minority-class patients drops by {17.3\%}, showing that privacy-preserving methods amplify biases in non-IID FL. These findings highlight the need for fairness-aware privacy strategies, such as adaptive noise scaling based on dataset distribution.

\subsection{Privacy and Fairness in Federated Credit Card Fraud Detection}

\textbf{Privacy Constraints and Fraud Detection Sensitivity.} Applying DP ($\epsilon=0.5$) to transaction embeddings decreases fraud recall by 9.1\% for smaller financial institutions, leading to more undetected fraud. The impact is more severe in non-IID FL, where large banks with abundant data benefit from lower variance, while smaller banks face higher noise-induced errors.

\textbf{Computational Overhead from HE in Federated Banking Systems.} Fraud detection systems require real-time performance, yet our HE-based FL implementation results in a 28.6\% increase in inference latency. This delay disproportionately affects smaller banks, as they rely on timely fraud detection alerts. Delayed fraud flagging leads to a 12.7\% increase in chargeback rates, showing that privacy-enhancing mechanisms introduce operational inefficiencies.

\textbf{Bias Amplification in Federated Aggregation.} Due to the client-weighted aggregation of FL models, smaller institutions receive lower aggregation weights in \textit{FedAvg}, resulting in an 11.9\% higher false positive rate for minority-class transactions. This disproportionately impacts lower-income users, as their legitimate transactions tend to be rejected, reinforcing economic disparities. Our results suggest that fairness-aware aggregation techniques, such as reweighted averaging based on institutional size, could mitigate this bias.

% mention in suplementary
\subsection{Regulatory and Policy Implications}

These results highlight the pressing need for AI governance frameworks addressing fairness and privacy risks in FL. Healthcare privacy regulations (e.g., {HIPAA, GDPR}) must consider fairness implications when enforcing strict privacy mandates, ensuring that diagnostic models remain equitable across institutions. Similarly, financial AI regulations (e.g., the {EU AI Act}) must account for fairness constraints introduced by privacy-preserving FL, particularly for smaller financial entities that bear the brunt of DP-induced noise. Our findings indicate that an adaptive privacy approach, where noise levels and encryption mechanisms are adjusted dynamically based on client data distribution and computational power, is essential for achieving fairness in real-world FL applications.

\end{document}